\documentclass[journal,twoside,web]{ieeecolor}
\usepackage{generic}
\usepackage{cite}
\usepackage{amsmath,amssymb,amsfonts}
\usepackage{algorithmic}
\usepackage{graphicx}
\usepackage{textcomp}
\usepackage{authblk}
\usepackage[capitalize]{cleveref}
 \crefformat{appendix}{#2#1#3}
\usepackage{subcaption}
\usepackage{subfig}
\usepackage{booktabs}
\usepackage{tabularx}
\usepackage{multirow}
\usepackage{multicol}
\usepackage[ruled,vlined,linesnumbered]{algorithm2e}
\usepackage{xurl}
\usepackage{siunitx}
\usepackage{placeins}
\newcommand{\ra}[1]{\renewcommand{\arraystretch}{#1}}
\newcommand{\textscp}[1]{\textsc{#1}{\small s}}

\def\BibTeX{{\rm B\kern-.05em{\sc i\kern-.025em b}\kern-.08em
    T\kern-.1667em\lower.7ex\hbox{E}\kern-.125emX}}
\begin{document}
\title{Estimating Ground Reaction Forces \\ from Inertial Sensors}

\author{
    B. Song, 
    M. Paolieri,
    H. E. Stewart,
    L. Golubchik,
    J. L. McNitt-Gray,
    V. Misra,
    D. Shah
\thanks{B. Song, M. Paolieri, H. E. Stewart, L. Golubchik, and J. L. McNitt-Gray are with University of Southern California, Los Angeles, USA (e-mails: bowenson@usc.edu, paolieri@usc.edu, hestewar@usc.edu, leana@usc.edu, mcnitt@usc.edu). V. Misra is with Columbia University, New York, USA (e-mail: vishal.misra@columbia.edu). D. Shah is with Massachusetts Institute of Technology, Cambridge, USA (e-mail: devavrat@mit.edu). Part of this work was done while L. Golubchik was visiting Columbia University and MIT.}
\thanks{Submitted to IEEE Transactions on Biomedical Engineering on Jan. 8, 2024, accepted on Sep. 14, 2024. Copyright (c) 2024 IEEE. Personal use of this material is permitted. However, permission to use this material for any other purposes must be obtained from the IEEE by sending an email to pubs-permissions@ieee.org.}
}

\maketitle
\begin{abstract}
    \emph{Objective:}
Our aim is to determine if data collected with inertial measurement units (IMUs) during steady-state running could be used to estimate ground reaction forces (GRFs) and to derive biomechanical variables (e.g., contact time, impulse, change in velocity) using lightweight machine-learning approaches. In contrast, state-of-the-art estimation using LSTMs suffers from prohibitive inference times on edge devices, requires expensive training and hyperparameter optimization, and results in black box models.
\emph{Methods:}
We proposed a novel lightweight solution, SVD Embedding Regression (SER), using linear regression between SVD embeddings of IMU data and GRF data. We also compared lightweight solutions including SER and k-Nearest-Neighbors (KNN) regression with state-of-the-art LSTMs.
\emph{Results:}
We performed extensive experiments to evaluate these techniques under multiple scenarios and combinations of IMU signals and quantified estimation errors for predicting GRFs and biomechanical variables. We did this using training data from different athletes, from the same athlete, or both, and we explored the use of acceleration and angular velocity data from sensors at different locations (sacrum and shanks). 
\emph{Conclusion:}
Our results illustrated that lightweight solutions such as SER and KNN can be similarly accurate or more accurate than LSTMs. The use of personal data reduced estimation errors of all methods, particularly for most biomechanical variables (as compared to GRFs); moreover, this gain was more pronounced in the lightweight methods.
\emph{Significance:}
The study of GRFs is used to characterize the mechanical loading experienced by individuals in movements such as running, which is clinically applicable to identify athletes at risk for stress-related injuries.

\end{abstract}


\newcommand{\SER}{\textit{SVD-based output-Embedding Regression}}

\section{Introduction\label{sec:intro}}

Ground reaction force (GRF), the force exerted by the ground on a body during contact, is a key measurement used in biomechanics to study the whole body dynamics of human movement. It characterizes the mechanical loading of the body, which contributes to the stress response of bone and soft tissue~\cite{Munro87}.
Analysis of GRFs has been proposed as a means to identify factors that lead to bone stress injuries for runners~\cite{Cavanagh80,James78,Hreljac04,Napier18,johnson2020impact,vannatta2020biomechanical}.
%
Determining the cause of running-related injuries continues to be challenging in part because of the inability to account for the mechanical loading experienced by an individual during multiple foot contacts within and across training sessions.
Despite conflicting findings~\cite{matijevich2019ground}, recent studies \cite{rice2024speed} continue to explore the role of GRFs in identifying mechanisms contributing to lower extremity injuries.
Understanding GRF characteristics is important for improving performance (e.g., greater net impulse translates to greater changes in body momentum) and may provide insights into mechanisms like bending moments imposed on the lower extremity during ground contact, which could help further identify combinations of factors leading to injury.
To this end, domain experts find that the GRF waveform and the biomechanical variables derived from it, such as contact time, impulse, and change in center of mass velocities provide meaningful context for understanding how GRFs cause observed body movements and contribute to stress responses~\cite{Bigouette16,Kiernan18,Napier18,Messier18}.

Direct measurement of GRFs is typically performed using force plates or instrumented treadmills in a laboratory environment~\cite{Riley08, Kluitenberg12,Asmussen19}. Wearable GRF sensors have been proposed~\cite{Jacobs15} but there is a lack of reliable sensors on the market. For instance, authors in \cite{mason2023wearables} note that overall validity and reliability of these devices appears to be system, location, and speed dependent.
Machine learning methods can mitigate the challenges and costs associated with direct data collection by estimating signals of interest from other signals more readily accessible and from cost-effective sources.


Our study evaluates machine learning approaches to estimate GRFs from inertial measurement unit (IMU) signals collected on regular treadmills. Given the high cost of instrumented treadmills, this approach can provide more people with valuable biomechanical data about their performance at a much lower cost, thus enabling similar studies with more frequent data collections while lowering barriers for athletes, coaches, and researchers.
This data can help elucidate the relationship between GRFs and performance to advance the overall understanding of biomechanics and potential injury interventions.

The estimation of GRFs from IMU sensor data is considered in previous works \cite{leporace2018prediction,Dorschky20,Johnson21,Alcantara21,Alcantara22} to overcome the difficulty of direct GRF measurement.
State-of-the-art approaches use deep learning to estimate GRFs; for example, \cite{Dorschky20} uses convolutional neural networks to estimate GRFs from acceleration and angular velocity waveforms (collected using low-cost wearable IMUs), while \cite{Alcantara22} uses LSTM neural networks to estimate GRFs from acceleration waveforms (collected on regular treadmills).
Drawbacks of these approaches include their often  prohibitive inference times on edge devices, requirement of expensive resources to support long training times and hyperparameter optimization, and the black box nature of the resulting models, which provide limited insights for the study of the relationship between IMUs and GRFs, or to identify and remove bad training data (this is defined by provenance as explained in \cref{sec:provenance}).

In this work, we address these limitations by exploring lightweight alternatives to deep-learning methods to facilitate training and inference on devices with limited computing power. We also analyze the improvements resulting from the use of training data collected for the target athletes at multiple body locations.
Specifically, we compare state-of-the-art approaches based on LSTM neural networks with two lightweight approaches: \emph{SVD Embedding Regression} (SER), our proposed approach to estimate GRFs through linear regression between singular value decomposition (SVD) embeddings of IMU data (input) and GRF data (output); and \emph{$k$-Nearest Neighbors} (KNN) regression.
We evaluate the errors of these techniques in estimating GRFs and derived biomechanical variables when using training data collected (1)~from different athletes, (2)~from the same athlete, (3)~or both.
In each scenario, we explore the use of \emph{acceleration and angular velocity data} from sensors positioned at \emph{different locations} (sacrum, left and right shanks which are positioned directly above the left and right lateral malleolus). This data can be easily collected given the wide availability of wearable IMUs measuring linear acceleration, angular velocity, and magnetic fields concurrently.

To evaluate the efficacy of different machine learning methods under a variety of scenarios and input sensors, we use an existing set of deidentified data collected by domain experts working with collegiate distance runners in the NCAA Pac-12 conference. Details on the data collection and preprocessing are described in \cref{sec:dataset}, while the different estimation tasks and metrics are defined in \cref{sec:tasks}.

Our work provides the following contributions.
\begin{enumerate}
    \item We propose \textsc{ser}, a novel approach to estimate GRFs from IMU measurements (\cref{sec:ser}), as an alternative to \textsc{knn} (\cref{sec:knn}) and \textsc{lstm} neural networks (\cref{sec:RNN}).
    \item Through our experimental results (\cref{sec:results}), we show that simple machine learning methods such as \textsc{ser} and \textsc{knn} can be similarly accurate or more accurate than \textsc{lstm} neural networks, requiring fewer computing resources and energy, while allowing much faster training times and hyperparameter optimization.
    \item By carrying out the evaluation of all machine learning methods in all scenarios using only acceleration, only angular velocity, or both, we allow a direct comparison on the same dataset and show that angular velocity measurements (collected by IMU sensors) reduce GRFs estimation error when combined with acceleration measurements.
    \item We show that GRF estimation error is reduced when using sensors at both sacrum and shanks, especially with \textsc{lstm}.
    \item We illustrate how personal training data  significantly reduces GRF estimation error of \textsc{knn} and \textsc{ser} for all combinations of input sensors.
\end{enumerate}

Notably, when personal training data are available, SER and kNN achieve rRMSE lower than 5\% for vertical GRF; as a reference, 6\% rRMSE is considered to be acceptable for the study of whole body dynamics in \cite{ren2008whole}.

\section{Notation and Dataset\label{sec:dataset}}

\begin{figure}[ht]
    \centering
    \includegraphics[width=\linewidth]{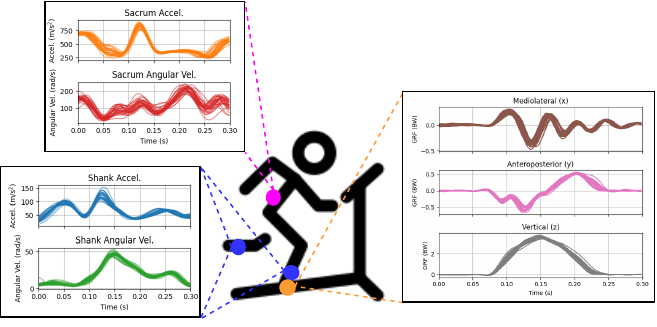}
    \caption{Example of signals measured during running from the respective sensors. (Right) 3 components of GRFs, i.e., the $x,y,z$ axes (mediolateral, anterior-posterior, vertical, respectively), measured from instrumented treadmills. (Left, Top) magnitude of IMU signals including acceleration and angular velocity from the sacrum. (Left, Bottom) magnitude of IMU signal acceleration and angular velocity from the left and right shanks}
    \label{fig:experiment_setup}
\end{figure}

We use an existing Pac-12 dataset consisting of 44 competitive collegiate distance runners (25 female and 19 male) from University of Colorado Boulder, University of Oregon, and Stanford University in accordance with the Institutional Review Board for research involving human participants.\footnote{This study is approved by the University of Oregon on April 27, 2021, protocol number 05162017.019. A subset of this dataset is used in \cite{harper2023,Alcantara21}.}
The dataset includes 114 collections (1--6 per participant) where participants ran on instrumented treadmills at multiple speeds: male participants ran at 7, 6.5, and 5 min/mi (3.8, 4.1, 5.4 m/s) and female participants ran at 7 and 5.5 min/mi (3.8, 4.9 m/s); GRF data was collected by the treadmills at 1,000 Hz, while wearable IMU sensors collected acceleration and angular velocity data at the sacrum, left shank, and right shank with 500 Hz frequency.
Including all athletes, collections, and running speeds, the dataset provides 276 running intervals, each with at least 60 steps (approximately 15 seconds).  Examples of our collected signals are shown in \cref{fig:experiment_setup}.

We synchronize data using events from various sensors to divide each running interval into individual foot contacts. To minimize noise commonly found in IMU and GRF data~\cite{Alcantara22}, we employ a 4th order Butterworth low-pass filter. This filter has a cutoff frequency of 20~Hz for acceleration and angular velocity signals, and 30~Hz for GRF signals.
Applying the same filters allows us to make a fair quantitative comparison with related work. A detailed description of data preprocessing is provided in Appendix \ref{app:manual_alignment}.

The integration of the data from all athletes and their data collections at different speeds resulted in a dataset of 16,000 steps (after splitting the IMU/GRF signals using 400 ms windows aligned using cross-correlation). Each step is characterized by 15 time-series signals including the 3 components ($x, y, z$) of the GRFs~$\vec{g}(t)$, sacrum acceleration~$\vec{a}_s(t)$, sacrum angular velocity~$\vec{\omega}_s(t)$, left or right shank acceleration~$\vec{a}_{lr}(t)$, and left/right shank angular velocity~$\vec{\omega}_{lr}(t)$. Each signal is sampled at 500 Hz with a fixed 400 ms time window (which results in 200 time points per window and a single step per window).

\section{Estimation Tasks and Metrics\label{sec:tasks}}

\subsection{Estimation Tasks and Hyperparameter Selection}
We consider different data scenarios for the estimation of GRF data: similarly to related work \cite{Alcantara22} using leave-one-subject-out for evaluation, we estimate GRFs of a target athlete using subject-independent training data collected only from other athletes (we refer to this scenario as ``\textsc{others}''); in addition, we consider subject-dependent scenarios where training data from the target athlete is used exclusively (scenario ``\textsc{personal}'') or in conjunction with data from other athletes (scenario ``\textsc{everyone}''). For each scenario, we consider different input signal cases to estimate GRFs as listed in \cref{tab:input-acronyms}.

\begin{table*}[ht]\centering
    \ra{1.3}
    \begin{tabular}{@{}lll@{}}\toprule
    Acronym & Input Signals & Description\\\midrule
    \textsc{all} & $\lVert\vec{a}_s(t)\rVert,\lVert\vec{a}_{lr}(t)\rVert,\lVert\vec{\omega}_s(t)\rVert,\lVert\vec{\omega}_{lr}(t)\rVert$ & L2 norm of all acceleration and angular velocity signals\\
    \textsc{acc} & $\lVert\vec{a}_s(t)\rVert,\lVert\vec{a}_{lr}(t)\rVert$ & L2 norm of acceleration signals (sacrum, left/right shanks)\\
    \textsc{ang} & $\lVert\vec{\omega}_s(t)\rVert,\lVert\vec{\omega}_{lr}(t)\rVert$ & L2 norm of angular velocity signals (sacrum, left/right shanks)\\
    \textsc{sacrum} & $\lVert\vec{a}_s(t)\rVert,\lVert\vec{\omega}_s(t)\rVert$ & L2 norm of acceleration and angular velocity at the sacrum\\
    \textsc{shanks} & $\lVert\vec{a}_{lr}(t)\rVert,\lVert\vec{\omega}_{lr}(t)\rVert$ & L2 norm of acceleration and angular velocity at left/right shanks\\
    \textsc{sac/acc3d} & $\vec{a}_s(t)$ & $x,y,z$ components of acceleration signal at the sacrum\\
    \textsc{sac/acc} & $\lVert\vec{a}_s(t)\rVert$ & L2 norm of acceleration signal at the sacrum\\
    \bottomrule
    \end{tabular}
    \caption{Combinations of input signals for the estimation tasks}\label{tab:input-acronyms}
\end{table*}

Data scenarios, input signal cases, and machine learning methods are used to define \emph{estimation tasks}, each identified by a tuple $(\textit{scenario}, \textit{sensors}, \textit{method})$ with $\textit{scenario}\in\{\textsc{others}, \textsc{personal}, \textsc{everyone}\}$, $\textit{sensors}\in\{\textsc{all}, \textsc{acc}, \textsc{ang}, \textsc{sacrum}, \textsc{shanks}, \textsc{sac/acc3d}, \textsc{sac/acc}\}$, and $\textit{method}\in\{\textsc{ser},\textsc{knn},\textsc{lstm}\}$.

To allow error comparisons across estimation tasks, we select 10 athletes with the largest amount of data and use their last collections (including multiple running speeds) as test data.
Specifically, let $\textit{Test}_i, i=1,\dots,10$ represent the last collections of these athletes, $\textit{Train}_i$ their other collections, and $\textit{Train}_{\textit{REST}}$ the data of the remaining athletes.

\begin{itemize}
\item In the \textsc{others} scenario, we leave out one of the $\textit{Test}_i, i=1,\dots,10$ as testing set and use all $\textit{Train}_j$ and $\textit{Test}_j$ with $j\neq i$, and $\textit{Train}_{\textit{REST}}$ as training set to select hyperparameters (with $k$-fold cross validation; each fold has the data of 8 of the 43 remaining athletes), resulting in 10 different models.
\item In the \textsc{personal} scenario, we leave out one of the $\textit{Test}_i, i=1,\dots,10$ as testing set and use only $\textit{Train}_i$ as training set to select hyperparameters (with $k$-fold cross validation; each fold has the data of 1 collection of the athlete), also resulting in 10 different models.
\item In the \textsc{everyone} scenario, we leave out all of $\textit{Test}_i,i=1,\dots,10$ as testing set and use all of $\textit{Train}_i,i=1,\dots,10$ and $\textit{Train}_{\textit{REST}}$ as training set to select hyperparameters (with $k$-fold cross validation; each fold has the data of 20 collections), resulting in a single model.
\end{itemize}

For each estimation task, after hyperparameter optimization with $k$-fold cross validation, the entire training set is used for training. Reported estimation error is the average obtained by models $i=1,\dots,10$ on $\textit{Test}_i$ in the \textsc{others} and \textsc{personal} scenarios, or by the single model of \textsc{everyone} on $\textit{Test}_i,i=1,\dots,10$. Note that test data of a model is never used for training nor hyperparameter selection; data with multiple running speeds is included during training and hyperparameter optimization, and also during testing.
For \textsc{lstm} training and hyperparameter optimization, we use early stopping with 30-epoch patience (i.e., select the best parameters observed during training, which continues until no improvements in validation error are observed for 30 epochs).

We also present results for an additional scenario of interest for the \textsc{lstm} method, where hyperparameter selection and training are carried out as in the \textsc{others} scenario, but the resulting models are then fine-tuned using $\textit{Train}_i$ (as in the \textsc{personal} scenario, using $k$-fold cross validation to select the number of fine-tuning epochs) before evaluating their estimation error on $\textit{Test}_i$; this scenario, which results in a model for each athlete, is particularly common for neural networks with many parameters, where using a pretrained model mitigates the issue of data scarcity. 

\subsection{Error Metrics for Estimated GRF Waveforms}

The proposed machine learning methods estimate, for each step, the components of the GRFs $\vec{g}(t) = \big(g_x(t), g_y(t), g_z(t)\big)$ at each time point $t=1,\dots,T$. We indicate the estimations by $\hat g_x(t)$, $\hat g_y(t)$, and $\hat g_z(t)$, respectively, and we evaluate the Root Mean Squared Error (RMSE) and the Relative Root Mean Squared Error (rRMSE) for each component $d\in\{x,y,z\}$:
\begin{align}
    RMSE(g_d,\hat{g}_d) &= \sqrt{\frac{\sum_{t=1}^T [g_d(t)-\hat{g}_d(t)]^2}{T}}\label{eq:rmse}\\
    rRMSE(g_d,\hat{g}_d) &= \frac{RMSE(g_d,\hat{g}_d)}{[\textsc{range}(g_d) + \textsc{range}(\hat g_d)]/2}\label{eq:rrmse}
\end{align}
where $\textsc{range}(v) = \max_{t=1,\dots,T} v(t) - \min_{t=1,\dots,T} v(t)$. We compute the average of these error metrics across estimated steps.  For reference, related literature has reported $RMSE(g_z, \hat g_z)$ in the range of $0.14$~BW (i.e., N/kg) to $0.21$~BW~\cite{komaris2019predicting,Dorschky20} for vertical GRF estimations and $rRMSE(g_z, \hat g_z)$ in the range of $6\%$ to $14\%$ \cite{ren2008whole,Johnson21}; we note that the differences in results are partly due to differences in participants, sensors, and data curation between the datasets used in related works.
Note also that GRFs are normalized by body weight in our dataset; even when omitted, RMSE errors are relative to body weights of the athletes.

\subsection{Error Metrics for Estimated Biomechanical Variables}
\label{sec:gait_metrics}

GRF waveforms measured during foot contact are frequently used by domain experts to calculate discrete biomechanical variables representing different characteristics of a running step. We consider the biomechanical variables \textit{Loading Rate, Contact Time, Braking Time, Braking Percentage, Active Peak, Average Vertical Force, Vertical Impulse, \text{and} A/P Velocity Change}
(defined in ~\cite{Munro87} and Appendix \ref{app:gait_metrics}).
We evaluate each biomechanical variable $f$ from the estimated GRF waveforms $\hat g_x(t), \hat g_y(t), \hat g_z(t)$ and from their actual values $g_x(t), g_y(t), g_z(t)$ for $t=1,\dots,T$, and we compute the mean absolute percentage error (MAPE), i.e., the mean of $|f(\hat{g}_x, \hat{g}_y, \hat{g}_z) - f(g_x, g_y, g_z)|/|f(g_x, g_y, g_z)|$ across different steps. In addition to these metrics, we also study the effects of different estimation models on the resulting waveforms and their interpretability in \cref{sec:results}.

\section{Methods\label{sec:methods}}

\subsection{SVD Embedding Regression (SER)\label{sec:ser}}
As a lightweight alternative to deep learning methods for the estimation of GRFs from IMU signals (acceleration and angular velocity at different body locations), we propose the use of linear regression between SVD embeddings of input (IMU) and output (GRF) data; to reconstruct the GRF signals from the predicted output embedding (the \emph{pre-image problem}), we use the right singular vectors of the training data.
This approach (which can be viewed as a natural generalization of \emph{Principal Component Regression} at a high dimension~\cite{bair2006prediction}) is similar to transduction of structured data~\cite{Cortes05}, but pre-image calculation is very fast, providing a lightweight alternative to deep learning methods.

\subsubsection{SVD Embedding of IMU and GRF Signals\label{sec:svd}}

\begin{figure*}[ht]
    \centering
    \begin{subfigure}[b]{.57\linewidth}
        \includegraphics[width=\linewidth]{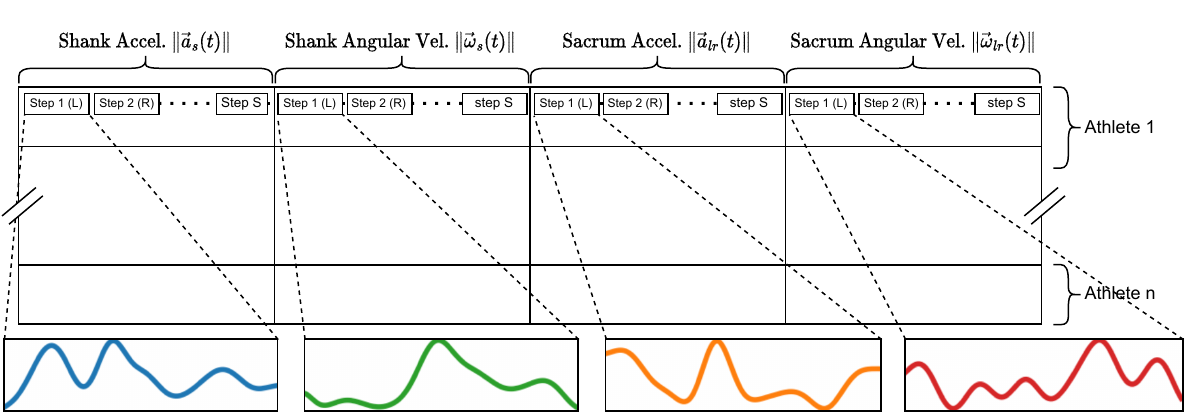}
        \caption{$A_{IMU}$\protect\label{fig:imu_mtx}}
    \end{subfigure}
    \hfill
    \begin{subfigure}[b]{.42\linewidth}
        \includegraphics[width=\linewidth]{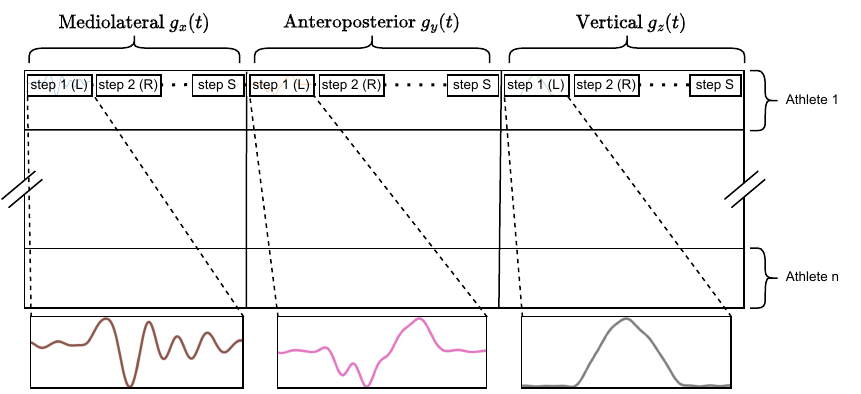}
        \caption{$A_{GRF}$\protect\label{fig:grf_mtx}}
    \end{subfigure}
    \caption{Organization of input (IMU) and output (GRF) matrices for the SER method. Each row includes data for multiple left/right steps and multiple signals; data of an athlete can span multiple rows.}
    \label{fig:mtx_arrangement}
\end{figure*}

We organize our training data into two matrices, $A_{IMU}$~\cref{fig:imu_mtx} and $A_{GRF}$~\cref{fig:grf_mtx}. Each row of these matrices corresponds to a different batch of $S$ consecutive running foot contacts from a measurement (i.e., steps of an athlete running at a given speed); for each running step in a batch and time point~$t$ (200~per step), the columns of $A_{IMU}$ include the IMU signals (e.g., the L2~norm of acceleration and angular velocity signals $\lVert\vec{a}_s(t)\rVert,\lVert\vec{a}_{lr}(t)\rVert,\lVert\vec{\omega}_s(t)\rVert,\lVert\vec{\omega}_{lr}(t)\rVert$ in the \textsc{all} case), while the columns of $A_{GRF}$ include the components of the GRFs, i.e., $g_x(t), g_y(t), g_z(t)$.

To obtain low-dimensional embeddings of the training data, we compute the SVD decomposition of the matrices $A_{IMU} \in \mathbb{R}^{n\times m}$ and $A_{GRF} \in \mathbb{R}^{n\times p}$, i.e.,
\begin{equation}\label{eq:svd}
    \begin{split}
    A_{IMU} &= U_{IMU} \Sigma_{IMU} V_{IMU}^T\\
    A_{GRF} &= U_{GRF} \Sigma_{GRF} V_{GRF}^T
    \end{split}
\end{equation}
where: $U_{IMU} \in \mathbb{R}^{n\times n}$ and $U_{GRF} \in \mathbb{R}^{n\times n}$ are orthogonal matrices (with left singular vectors as columns); $\Sigma_{IMU} \in \mathbb{R}^{n\times m}$ and $\Sigma_{GRF} \in \mathbb{R}^{n\times p}$ are rectangular diagonal matrices (with singular values in ascending order on the diagonal); $V_{IMU} \in \mathbb{R}^{m\times m}$ and $V_{GRF} \in \mathbb{R}^{p\times p}$ are orthogonal matrices (with right singular vectors as columns).

\begin{figure}[t]
    \centering
    \begin{subfigure}[b]{\linewidth}
        \includegraphics[width=\textwidth]{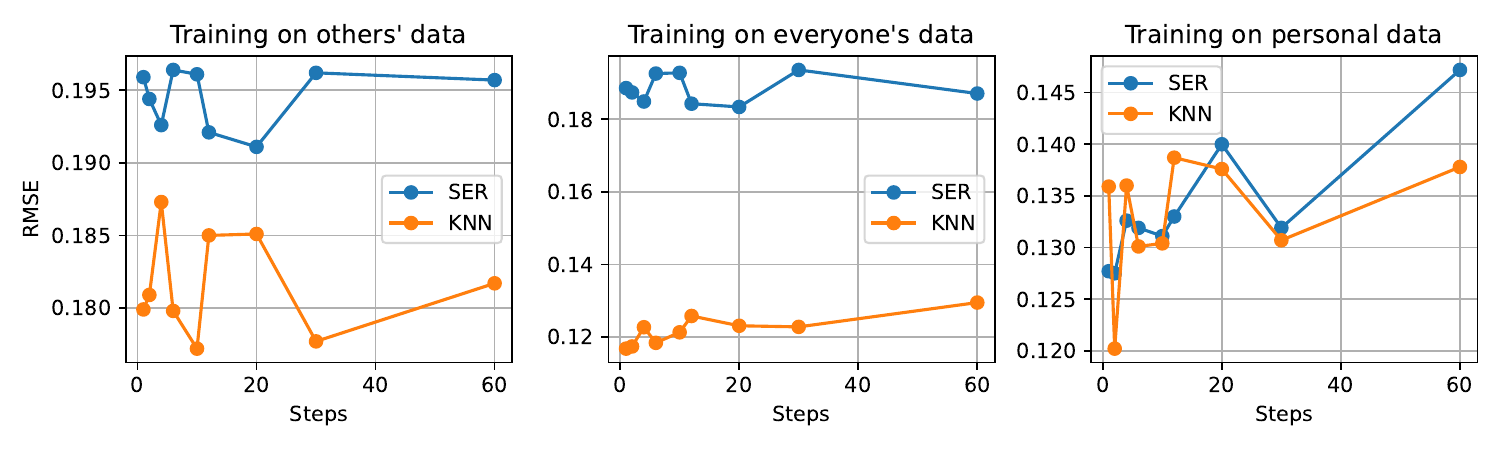}    \caption{Input from \textsc{all} signals}
    \end{subfigure}
    \begin{subfigure}[b]{\linewidth}
        \includegraphics[width=\textwidth]{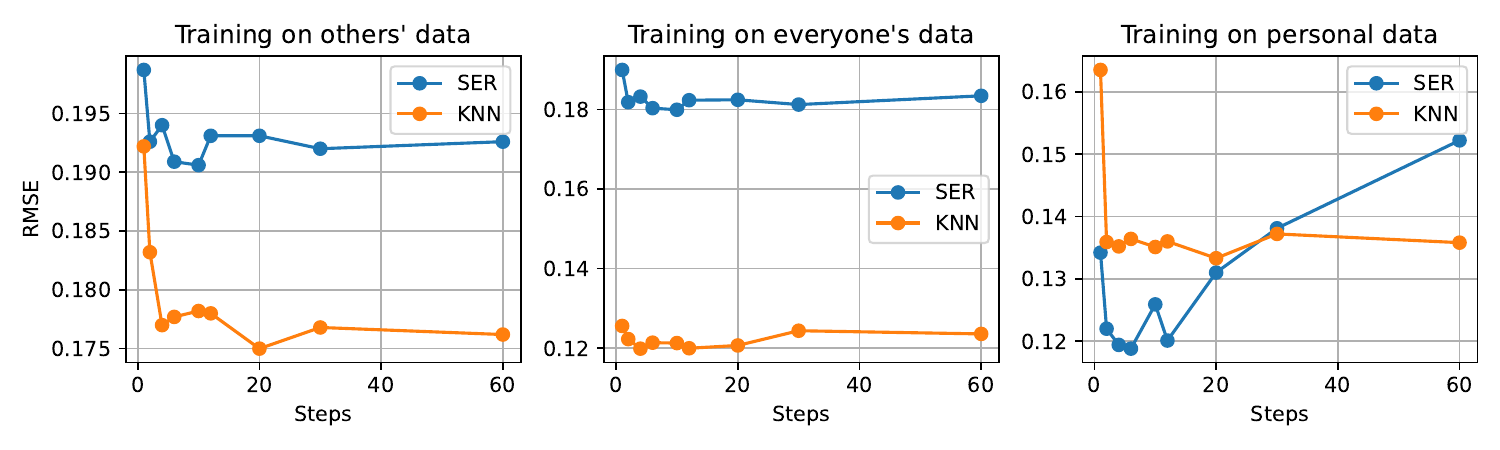}    \caption{Input from \textsc{sac/acc3d} signals}
    \end{subfigure}
    \caption{RMSE of vertical GRF estimated with SER and KNN for different numbers of steps $S$ in a batch}
    \label{fig:accuracy_vs_steps}
\end{figure}

We obtain low-rank approximations by keeping only the first $r$~singular values of the SVD decomposition, i.e., the first $r$~columns of the $U$ and $V$ matrices, and the first $r$~rows/columns of $\Sigma$:
\begin{equation}\label{eq:svd_approx}
    \begin{split}
    A_{IMU} &\approx \overline{U}_{IMU} \overline{\Sigma}_{IMU}\overline{V}_{IMU}^T\\
    A_{GRF} &\approx \overline{U}_{GRF} \overline{\Sigma}_{GRF} \overline{V}_{GRF}^T
    \end{split}
\end{equation}
with $\overline{U}_{IMU}\in \mathbb{R}^{n\times r_{IMU}}$, $\overline{\Sigma}_{IMU}\in \mathbb{R}^{r_{IMU}\times r_{IMU}}$, $\overline{V}_{IMU}\in \mathbb{R}^{m\times r_{IMU}}$ and $\overline{U}_{GRF}\in \mathbb{R}^{n\times r_{GRF}}$, $\overline{\Sigma}_{GRF}\in \mathbb{R}^{r_{GRF}\times r_{GRF}}$, $\overline{V}_{GRF}\in \mathbb{R}^{p\times r_{GRF}}$.
On our dataset, we use ranks $r_{IMU} = r_{GRF} = 6$, which retain at least $95\%$ of the energy of $\Sigma_{IMU}$ and $\Sigma_{GRF}$, respectively (i.e., the sum of the squares of the retained singular values is at least $95\%$ of the sum of the squares of all the singular values). The rank 6 approximation of GRFs has an average RMSE of 0.079 BW or rRMSE of 2.1\%; the same accumulative energy is also chosen in the literature \cite{venturi2023svd,liao2013efficient}. We also select the number of steps per row $S \in \{2, 3, 5, 6, 10, 12, 15, 20, 30, 60\}$, using a validation set.
Larger ranks $r_{IMU}$ and $r_{GRF}$ work similarly well, while the method is sensitive to $S$,
as illustrated in \cref{fig:accuracy_vs_steps}.

\subsubsection{Training and Estimation using SVD Embeddings}
\label{sec:grf_prediction_feature}

\begin{figure}[t]
    \centering
    \includegraphics[width=\linewidth]{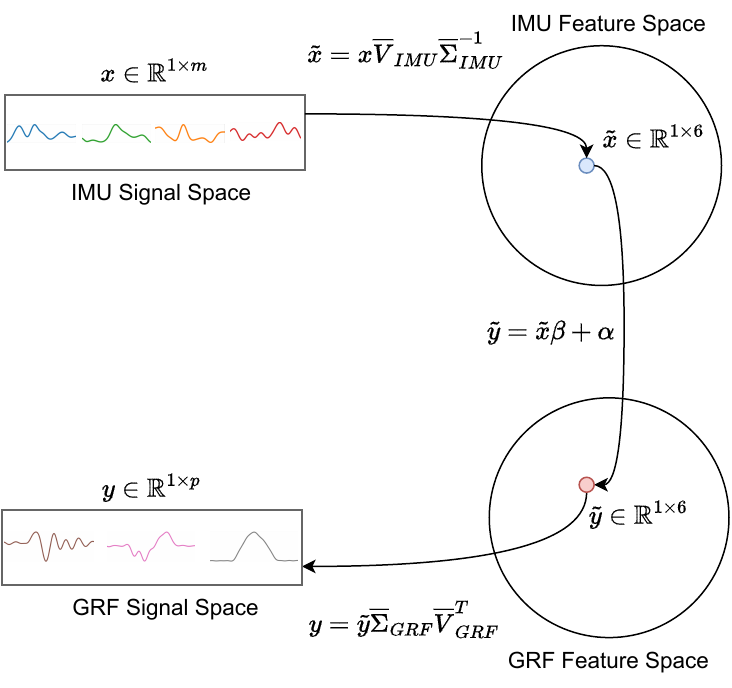}
    \caption{Estimating GRFs with \SER}
    \label{fig:model_prediction_diagram}
\end{figure}

After low-rank approximation, each row of matrices $\overline{U}_{IMU}$ and $\overline{U}_{GRF}$ is a vector with $r_{IMU}$ and $r_{GRF}$ components representing the embeddings (i.e., the features) of IMU (input) and GRF (output) signals, respectively, for a batch of $S$ running steps in the training set. We train a predictor for each component of the GRF embedding using \emph{least squares regression} with elastic net regularization, i.e., we select, for each $j=1,\dots,r_{GRF}$, the parameters $\beta_j \in \mathbb{R}^{r_{IMU}}$ and $\alpha_j \in \mathbb{R}$ minimizing the loss
\begin{align*}
\label{eq:feature_linear_regression}
     \sum_{i=1}^n & \left[(\overline{U}_{GRF})_{ij} - \left((\overline{U}_{IMU})_{i*} \beta_j + \alpha_j\right)\right]^2\ \\
     & + \lambda_2 \lVert \beta_j \rVert_2^2
     + \lambda_1 \lVert \beta_j \rVert_1
\end{align*}
where $(\overline{U}_{IMU})_{i*} \in \mathbb{R}^{r_{IMU}}$ represents the $i$th row of $\overline{U}_{IMU}$ and $(\overline{U}_{GRF})_{ij} \in \mathbb{R}$ represents component $j$ of the GRF embedding for the $i$th training example.
The regularization weights $\lambda_1 \geq 0$ and $\lambda_2 \geq 0$ are selected for each estimation task using a validation set.

Given the IMU signals $x \in \mathbb{R}^{1\times m}$ of a new sequence of $S$~steps, we estimate the GRF signals $y \in \mathbb{R}^{1\times p}$ by (\cref{fig:model_prediction_diagram}):
\begin{enumerate}
    \item Calculating the embedding $\tilde x \in \mathbb{R}^{1\times r_{IMU}}$ of the new IMU signals as $\tilde x = x \overline{V}_{IMU} \overline{\Sigma}_{IMU}^{-1}$;
    \item Predicting the embedding $\tilde y \in \mathbb{R}^{1\times r_{GRF}}$ of the corresponding GRF signals as $(\tilde y)_j = \tilde x \beta_j + \alpha_j$ for each $j=1,\dots,r_{GRF}$;
    \item Reconstructing the estimated GRF signals $y \in \mathbb{R}^{1\times p}$ as $y = \tilde y \overline{\Sigma}_{GRF} \overline{V}_{GRF}^T$.
\end{enumerate}

\subsection{$k$-Nearest Neighbors Regression\label{sec:knn}}

\begin{figure}[t]
    \centering
    \includegraphics[width=0.8\linewidth]{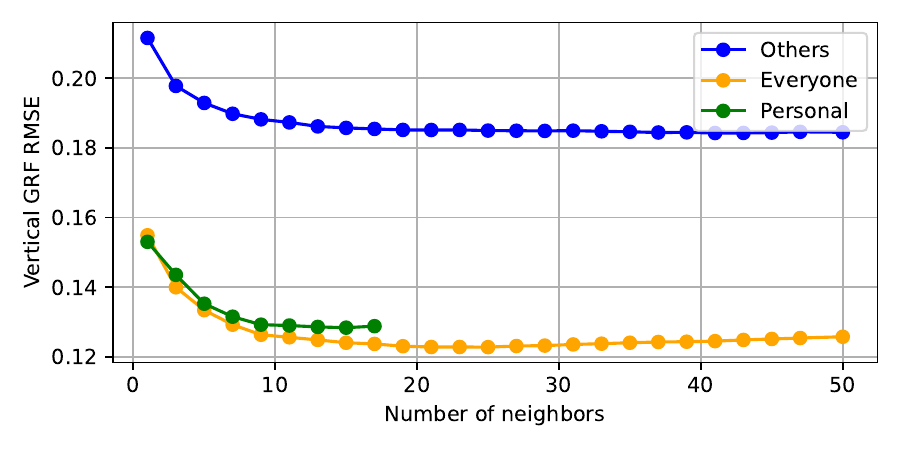}
    \caption{RMSE of vertical GRF for different numbers of neighbors $k$ using KNN regression}\label{fig:rmse_vs_k}
\end{figure}

As a lightweight baseline, we apply $k$-Nearest Neighbors regression (KNN), where the GRFs are estimated by combining the GRF signals of the $k$ training examples with most similar IMU signals.
Specifically, given the IMU signals $x \in \mathbb{R}^{1\times m}$ for a new sequence of $S$ steps, we estimate its GRF signals $y \in \mathbb{R}^{1\times p}$ by:
\begin{enumerate}
    \item Sorting the sequences of $S$ steps of the training set, $x_i \in \mathbb{R}^{1\times m}$ for $i=1,\dots, n$, by their Euclidean distances $d(x, x_i) = \lVert x - x_i \rVert_2$;
    \item Selecting the indices $\mathcal K \subseteq \{1,\dots,n\}$ of the $k$ training sequences with lowest distances;
    \item Estimating $y \in \mathbb{R}^{1\times p}$ as
    \[
        y = \frac{\sum_{i\in\mathcal K} d(x, x_i) \, y_i}{\sum_{i\in\mathcal K} d(x, x_i)}
    \]
    where $y_i\in \mathbb{R}^{1\times p}$ are the GRF signals associated with the IMU signals $x_i$ in the training set, for $i=1,\dots,n$.
\end{enumerate}

For each estimation task, we select the number of neighbors $k$ and the number of consecutive steps $S$ in a sequence using a validation set. While $k$ has a minor effect on estimation error (additional neighbors after $k=10$ have lower weights and provide minor improvements, as illustrated in \cref{fig:rmse_vs_k}), $S$ can have an important effect for some estimation tasks, as illustrated in \cref{fig:accuracy_vs_steps}.

\subsection{Long Short-term Memory Networks\label{sec:RNN}}

\begin{figure}[ht]
    \centering
    \includegraphics[width=\linewidth]{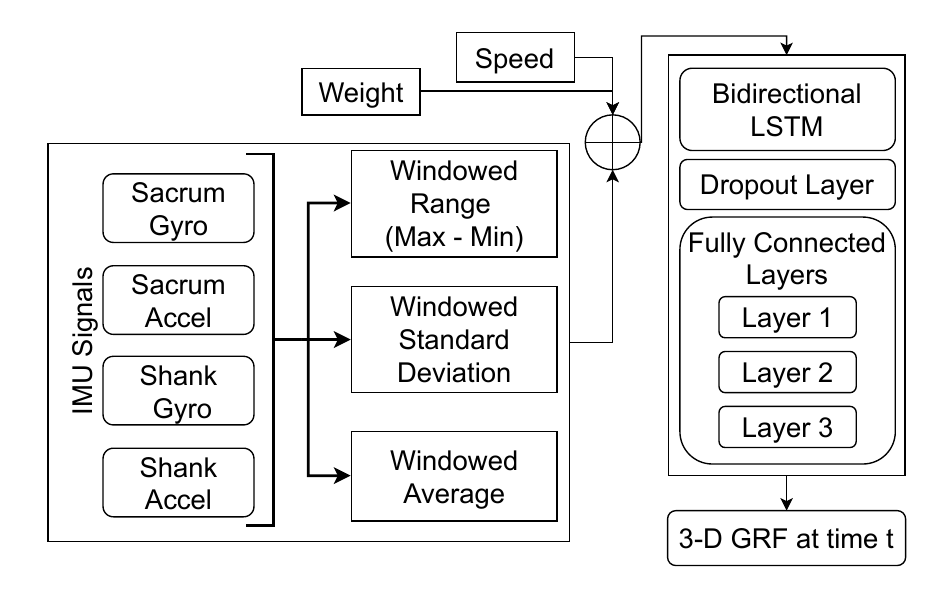}
    \caption{LSTM model architecture. The input are mean, standard deviation, and range calculated from IMU signals (\textsc{all} as shown in the figure)  for a time window, in addition to the weight and running speed of the athlete; a bidirectional LSTM layer is followed by dropout regularization and three fully connected layers.}
    \label{fig:rnn_network}
\end{figure}

\begin{table}[ht]
    \centering
    \begin{tabular}{@{}ll@{}}
    \toprule
    Hyperparameter & Values\\
    \midrule
    LSTM Units &  125, 250, 300, 500, 800, 1000 \\
    Layer 1 Neurons &  5, 10, 20, 35, 50\\
    Layer 2 Neurons &  10, 20, 35, 50 \\
    Layer 3 Neurons &  5, 10, 20, 35, 50 \\
    Batch Size &  8, 16, 32 \\
    Learning Rate &  0.0001, 0.0003, 0.0005, 0.0007, 0.001 \\
    Dropout Rate &  0.2, 0.4 \\
    \bottomrule
    \end{tabular}
    \caption{LSTM Hyperparameter Search Space}
    \label{tab:LSTM_search_space}
\end{table}

As a deep learning baseline, we adapt the state-of-the-art \textit{long short-term memory} (LSTM) neural network of \cite{Alcantara22} to estimate \emph{all components} of the GRFs (instead of the only vertical component estimated in \cite{Alcantara22}).
The model estimates $\vec{g}(t) = \big(g_x(t), g_y(t), g_z(t)\big)$ from the IMU signals in the time window $[t-W, t]$ of size $W>0$ (e.g., in the \textsc{all} case, $\lVert\vec{a}_s(u)\rVert,\lVert\vec{a}_{lr}(u)\rVert,\lVert\vec{\omega}_s(u)\rVert,\lVert\vec{\omega}_{lr}(u)\rVert$ for $u \in [t-W, t]$).
Similarly to \cite{Alcantara22}, we also provide the mean, standard deviation and range of IMU signals over the window $[t-W, t]$ as inputs, together with the running speed and the weight of the athlete.

The architecture of the network is depicted in \cref{fig:rnn_network}: after a bidirectional LSTM layer with \emph{tanh} activations and dropout (a regularization approach), we use three fully-connected layers with ReLU activations.
The model is trained using the \textit{Adam} optimizer as a standard choice and the mean square error as loss function.
We select the number of units of each layer, batch size, learning rate, and dropout rate using a validation set; the search space is reported in \cref{tab:LSTM_search_space}.
Notably, this machine learning approach has much higher training times than \textsc{ser} and \textsc{knn} (up to 1 hour for each combination of hyperparameters) and the largest search space (18,000 combinations of hyperparameters).
Due to memory limitations of the GPUs used for training (NVIDIA Titan X, 12 GB of RAM~\cite{nvidia_titan_gpu}), we were not able to train models with more than 1000 LSTM units, although hyperparameter selection suggests that a higher number of LSTM units could be beneficial, although quite costly.

\section{Results and Discussion\label{sec:results}}

\newcommand{\good}[1]{\color{blue}#1}
\newcommand{\bad}[1]{#1}
\begin{table*}[ht]\centering
    \ra{1.3}
    \begin{tabular}{@{}rcccccccccccc@{}}\toprule
        & \multicolumn{3}{c}{Scenario \textsc{others}} && \multicolumn{3}{c}{Scenario \textsc{everyone}} && \multicolumn{3}{c}{Scenario \textsc{personal}}\\\cmidrule{2-4}\cmidrule{6-8}\cmidrule{10-12}
        Input Signals & \textsc{ser} & \textsc{knn} & \textsc{lstm} && \textsc{ser} & \textsc{knn} & \textsc{lstm (fine-tuned)}         && \textsc{ser} & \textsc{knn} & \textsc{lstm} \\\midrule
        \textsc{all}        & 0.197        & 0.180          & \good{0.126} && 0.187 & \good{0.118} & \good{0.124} (\good{0.117})   && \good{0.130} & \good{0.122} & 0.134\\
        \textsc{acc}        & 0.220        & 0.197          & \good{0.177} && 0.210 & \good{0.125} & 0.175 (0.151)        && \good{0.127} & \good{0.127} & 0.143\\
        \textsc{ang}        & 0.197        & \good{0.187}   & \good{0.183} && 0.190 & \good{0.130} & 0.183 (0.180)        && \good{0.133} & \good{0.132} & 0.171\\
        \textsc{shank}      & \good{0.215} & \good{0.210}   & \good{0.206} && 0.205 & \good{0.149} & 0.209 (0.190)        && \good{0.139} & \good{0.134} & 0.188\\
        \textsc{sacrum}     & \good{0.217} &  \good{0.210}  & 0.289        && 0.205 & \good{0.129} & 0.286 (0.270)        && \good{0.136} & \good{0.137} & 0.184\\
        \textsc{sac/acc3d}  & 0.194        & \good{0.181}   & \good{0.171} && 0.185 & \good{0.122} & 0.177 (0.160)        && \good{0.128} & \good{0.132} & 0.178\\
        \textsc{sac/acc}    & 0.198        & \good{0.190}   & \good{0.187} && 0.190 & \good{0.130} & 0.206 (0.177)        && \good{0.129} & \good{0.133} & 0.193\\
        \bottomrule
    \end{tabular}
    \caption{RMSE (in body weight units, BW, i.e., N/kg) of vertical GRF estimations $RMSE(g_z, \hat g_z)$ for different input signals, data scenarios, machine learning methods (results highlighted in blue are optimal or less than 0.010 from optimal for a scenario and set of input signals)}\label{tab:rmse_z}
\end{table*}

\begin{table*}[th]\centering
    \ra{1.3}
    \begin{tabular}{@{}r*{12}{S[table-format=1.1]}@{}}\toprule
        & \multicolumn{3}{c}{Scenario \textsc{others}} && \multicolumn{3}{c}{Scenario \textsc{everyone}} && \multicolumn{3}{c}{Scenario \textsc{personal}}\\\cmidrule{2-4}\cmidrule{6-8}\cmidrule{10-12}
        Input Signals & \textsc{ser} & \textsc{knn} & \textsc{lstm} && \textsc{ser} & \textsc{knn} & \textsc{lstm (fine-tuned)} && \textsc{ser} & \textsc{knn} & \textsc{lstm} \\\midrule
        \textsc{all} & 6.5 & 6.0 & \good{4.2} && 6.2 & \good{3.9} & \good{4.2}\; $ (\good{3.9})$ && \good{4.2} & \good{4.0} & \good{4.3} \\
        \textsc{acc} & 7.4 & 6.6 & \good{6.0} && 7.0 & \good{4.1} & 5.9\; $ (5.2)$ && \good{4.1} & \good{4.1} & 4.8 \\
        \textsc{ang} & \good{6.5} & \good{6.2} & \good{6.1} && 6.3 & \good{4.3} & 6.0\; $ (6.0)$ && \good{4.3} & \good{4.3} & 5.9 \\
        \textsc{shank} & \good{7.2} & \good{7.0} & \good{7.0} && 6.8 & \good{4.9} & 7.0\; $ (6.7)$ && \good{4.5} & \good{4.4} & 6.5 \\
        \textsc{sacrum} & \good{7.2} & \good{6.9} & 10.1 && 6.8 & \good{4.2} & 9.8\; $ (9.6)$ && \good{4.4} & \good{4.5} & 6.3 \\
        \textsc{sac/acc3d} & 6.4 & \good{6.0} & \good{5.8} && 6.2 & \good{4.0} & 5.9\; $ (5.4)$ && \good{4.1} & \good{4.3} & 6.0 \\
        \textsc{sac/acc} & \good{6.6} & \good{6.3} & \good{6.3} && 6.3 & \good{4.3} & 6.9\; $ (6.1)$ && \good{4.2} & \good{4.4} & 6.5 \\
        \bottomrule
    \end{tabular}
    \caption{rRMSE (\%) of vertical GRF estimations $rRMSE(g_z, \hat g_z)$ for different input signals, data scenarios, machine learning methods (results highlighted in blue are optimal or less than 0.5\% from optimal for a scenario and set of input signals)}\label{tab:rrmse_z}
\end{table*}

\begin{table*}[th]\centering
    \ra{1.3}
    \begin{tabular}{@{}rcccccccccccc@{}}\toprule
        & \multicolumn{3}{c}{Scenario \textsc{others}} && \multicolumn{3}{c}{Scenario \textsc{everyone}} && \multicolumn{3}{c}{Scenario \textsc{personal}}\\\cmidrule{2-4}\cmidrule{6-8}\cmidrule{10-12}
        Input Signals & \textsc{ser} & \textsc{knn} & \textsc{lstm} (GPU) && \textsc{ser} & \textsc{knn} & \textsc{lstm} (GPU) && \textsc{ser} & \textsc{knn} & \textsc{lstm} (GPU) \\\midrule
        \textsc{all}        & \good{0.001} & 2.256 & 4.221 (0.497) && \good{0.001}  & 2.088 & 3.219 (0.472) && \good{0.001} & 0.042 & 4.922 (0.558) \\
        \textsc{acc}        & \good{0.003} & 1.165 & 4.325 (0.521) && \good{0.002}  & 1.089 & 3.251 (0.473) && \good{0.001} & 0.022 & 5.139 (0.586) \\
        \textsc{ang}        & \good{0.004} & 1.161 & 4.605 (0.543) && \good{0.0004} & 1.064 & 3.180 (0.474) && \good{0.001} & 0.025 & 4.179 (0.516) \\
        \textsc{shank}      & \good{0.005} & 1.165 & 4.014 (0.503) && \good{0.002}  & 1.062 & 3.216 (0.476) && \good{0.001} & 0.021 & 4.739 (0.554) \\
        \textsc{sacrum}     & \good{0.004} & 1.156 & 4.557 (0.545) && \good{0.0004} & 1.067 & 3.295 (0.480) && \good{0.001} & 0.023 & 4.945 (0.563) \\
        \textsc{sac/acc3d}  & \good{0.001} & 1.716 & 3.132 (0.439) && \good{0.002}  & 1.574 & 5.160 (0.579) && \good{0.001} & 0.030 & 0.784 (0.248) \\
        \textsc{sac/acc}    & \good{0.005} & 0.518 & 3.221 (0.450) && \good{0.002}  & 0.411 & 3.153 (0.461) && \good{0.001} & 0.013 & 0.862 (0.272) \\
        \bottomrule
    \end{tabular}
    \caption{Average inference time (in seconds) to estimate 3D GRFs of a collection (120-180 steps) for an athlete. Inference time is measured on an Intel i7-6800K CPU; for  \textsc{lstm} we also report average inference time using a TITAN X GPU.}\label{tab:inference_time}
\end{table*}

\begin{table}\centering
    \ra{1.3}
    \begin{tabular}{@{}ll@{}}\toprule
    Acronym & Training Data of the Scenario\\\midrule
    \textsc{others} & IMU and GRF data of other athletes\\
    \textsc{personal} & IMU and GRF data of the same athlete\\
    \textsc{everyone} & IMU and GRF data of all athletes\\
    \bottomrule
    \end{tabular}
    \caption{Scenarios for the estimation tasks}\label{scenario-acronyms}
\end{table}
\subsection{Estimation of the Normal GRF Waveform}

First, we focus on the estimation of RMSE and rRMSE for the vertical GRF, i.e., $RMSE(g_z, \hat g_z)$ and $rRMSE(g_z, \hat g_z)$, obtained by different machine learning methods for each set of input signals and scenario. 
The vertical component of the GRF is particularly important for the study of stress response in bone and soft tissue \cite{Cavanagh80,James78} and provides the information used to compute several discrete biomechanical variables (discussed in \cref{sec:gait_metrics_accuracy}).
Results are reported in \cref{tab:rmse_z,tab:rrmse_z}, while input signals and scenarios of different estimation tasks are summarized in \cref{tab:input-acronyms,scenario-acronyms}, respectively.
Although limited to our dataset, this evaluation allows us to make the following observations regarding the use of machine learning methods for the estimation of GRFs from IMU signals.

\subsubsection{Use of Acceleration and Angular Velocity Signals}
We observe that, in all scenarios and for all machine learning methods, the use of both acceleration and angular velocity signals is preferable, since RMSE and rRMSE are similar or significantly lower.
In particular, using only acceleration (case \textsc{acc}) or angular velocity signals (case \textsc{ang}) is generally worse than using both (case \textsc{all}).
Since most IMU sensors collect both signals, these improvements can be obtained without additional costs of the data collection process, despite the focus of related work on the exclusive use of acceleration signals \cite{Alcantara22}.

\subsubsection{Sensor Locations}
We observe that, in all scenarios and for all machine learning methods, the use of sensors at both sacrum and left/right shank (case \textsc{all}) is preferable, since RMSE and rRMSE are either similar or significantly lower than when each sensor location is used exclusively (cases \textsc{sacrum} and \textsc{shank}, respectively).
While expected for deep learning methods (\textsc{lstm}), this result highlights that lightweight machine learning methods such as \textsc{ser} and \textsc{knn} can also provide effective estimation models for complex IMU data collected from multiple locations.
%

\subsubsection{Use of Multidimensional Acceleration at the Sacrum}

Since related literature \cite{Alcantara21,Alcantara22} focuses on acceleration signals collected using sensors located at the sacrum, we explore the benefits of a multidimensional acceleration signal $\vec{a}_s(t)$ (\textsc{sac/acc3d}) to estimate GRFs, instead of its L2 norm $\lVert\vec{a}_s(t)\rVert$ (case \textsc{sac/acc}).
We observe that, in all scenarios and for all machine learning methods, the use of a multidimensional acceleration signal at the sacrum is preferable to its L2 norm, since RMSE and rRMSE are either similar or significantly lower.
We also note that, when limited to the sacrum location, the use of multidimensional acceleration signals (\textsc{sac/acc3d}) is preferable to the use of the L2 norm of both acceleration and angular velocity (\textsc{sacrum}), and even to the use of the L2 norm of only acceleration signals at the sacrum and left/right shanks (\textsc{acc}).
Our experiments indicate that the increase in the model computation cost is negligible.

\subsubsection{Use of Lightweight Machine Learning Methods}

While state-of-the-art approaches \cite{Alcantara22} adopt deep learning methods based on \textsc{lstm} neural networks, we observe that lightweight approaches can provide similar or lower error of the estimated GRFs for specific scenarios and input signals.
In the scenario \textsc{everyone}, \textsc{knn} is preferable to \textsc{lstm} and \textsc{ser}, as it provides much lower RMSE and rRMSE for most combinations of input signals; in the scenario \textsc{personal}, both \textsc{ser} and \textsc{knn} are preferable to \textsc{lstm}, for all combinations of input signals.
Notably, in the setting of \cite{Alcantara22} (scenario \textsc{others}, signals \textsc{sac/acc3d}), \textsc{knn} and \textsc{lstm} perform very similarly on our dataset (rRMSE of $6.0\%$ and $5.8\%$, respectively; as a reference, \textsc{lstm} incurred an RMSE of $6.4\%$ in \cite{Alcantara22}).

We attribute the improved performance of \textsc{knn} in the scenarios \textsc{everyone} and \textsc{personal} to the inclusion of historical running data collected for the target athlete in the training set: \textsc{knn} is able to exploit the patterns specific to the target athlete, while ignoring data from other athletes;
in contrast, \textsc{lstm} obtains similar estimation error in the \textsc{others} and \textsc{everyone} scenarios, but obtains lower error when used with sensors at multiple body locations (this observation holds in the fine-tuned results). Finally, \textsc{ser} is able to model the GRFs of an athlete very accurately in the \textsc{personal} scenario. 
We observe that the estimation error of a machine learning method in a specific scenario depends on multiple factors, including the total amount of available data (which is significantly lower in the \textsc{personal} scenario, possibly resulting in higher estimation error), the lack of personal data (which may result in higher estimation error in the \textsc{others} scenario), and the ability of an algorithm to hone in on personal data when it is mixed with that of other athletes (as in the \textsc{everyone} case).

Note that our dataset is relatively small (as is common in this field) and not all gait patterns are well-represented; estimation error improvements from the inclusion of personal data indicate that linear models can achieve greater specializations, rather than a lack of generalization to the broader population. 

\subsubsection{Resource and Energy Utilization}
We observe that the upfront training cost of \textsc{lstm} models is more than three orders of magnitude higher than that of \textsc{ser}, which drastically reduces GPU hours, carbon footprint, and cloud or on-premise computing costs. In our study, training of \textsc{lstm} models (including architecture search and hyperparameter tuning) required parallel processing on 13 TITAN~X GPUs for 30 days, while training of \textsc{ser} models (including exhaustive hyperparameter search) required only 5 hours on a single commodity CPU. Our wall-plug measurements of power consumption indicated 200~W for each GPU (rated at 250~W by Nvidia~\cite{nvidia_titan_energy}; utilization was 90\%) and only 90~W for an Intel i7-6800K CPU (rated at 140~W by Intel~\cite{intel_cpu_energy}), resulting in an energy cost of 1.9 Million Watt-Hours for \textsc{lstm} compared to 450 Watt-Hours for \textsc{ser}, giving a reduction in energy consumption by a factor of 4,160. 
Moreover, there is a substantial dollar cost when training \textscp{lstm} in the cloud. We estimate approximately over \$9,000 to train the \textscp{lstm} (e.g., using an AWS g4dn.metal instance with an on-demand hourly rate of \$7.8~\cite{amz_gpu_renting} for 9,360~GPU hours) and only \$36 to train \textsc{ser} (e.g., using an AWS hpc7a.12xlarge instance with an on-demand hourly rate of \$7.20~\cite{amz_cpu_renting} for 5 hours).

At the same time, the estimation error improvements due to \textsc{lstm} are only significant in one scenario (2\% rRMSE improvement when training on all IMU signals in the \textsc{others} scenario, while in all other cases the rRMSE improvement is at most 0.6\%).
Therefore, although the initial training time is a one-time cost, given that \textsc{lstm} models provide limited benefits, their additional cost in terms of time, energy, and dollar expenditure is a significant drawback.

\subsubsection{Efficient Inference for Near Real-Time Applications}
Considering inference latency for batch estimation, \textsc{ser} has the shortest inference time in all scenarios (\cref{tab:inference_time}) using CPUs. The inference time of \textsc{lstm} (using CPU or GPU) depends on the size of the model (i.e., number of parameters, layers, etc), which is determined through hyper-parameter search, whereas \textsc{knn}'s inference time depends on the size of training dataset.

We note that, while \textsc{ser} does not offer real-time inference, after the initial delay for the collection of 2-5 running steps (1-2 seconds) for inference, \textsc{ser}'s 0.3 ms inference latency (for estimating 20 steps\footnote{We report the latency for 20 steps, to make a meaningful comparison to LSTM, which uses 20 steps as in related literature.}) is well-suited for near real-time analysis performed entirely on an edge device such as a mobile phone. This is in contrast with \textsc{lstm} model inference which would require approximately 15 seconds (to estimate 20 steps) on the same mobile device (during which time an athlete would run at least another 30 steps). In order to achieve near real-time analysis (with some initial delay), inference time needs to be less than the running time for the steps being analyzed, or the analysis needs to skip some steps during inference. Another approach might be to process batches of steps. However, due to memory constraints, \textsc{lstm} inference time for batch analysis grows linearly with the batch size. 
Yet another approach to reducing LSTM inference latency on mobile devices is to use quantization. Our experiments with quantization indicated a reduction in LSTM's inference time down to 4 seconds, which is still more than 4 orders of magnitude higher than the inference latency of SER \emph{without} quantization. However, this reduction in inference latency comes at the cost of an increase in RMSE; our experiments indicated an average of $\approx 10\%$ increase in RMSE, with a worst case increase of over $34\%$ in the case of using \textsc{sac/acc 3d} as input signal.
Thus, \textscp{lstm} are less suitable for near real-time analysis on edge devices. 

We note that, to achieve 0.5 or 4~seconds inference latency for LSTM reported in \cref{tab:inference_time} requires use of cloud GPUs or CPUs, respectively. Note also that on-device prediction facilitates preservation of data privacy. The reduced cost and energy consumption of simpler methods such as \textsc{ser} also facilitates development of GRF estimation methods on embedded components which can be integrated into treadmills in an inexpensive manner (e.g., using embedded CPUs). This opens the way to integration of estimation techniques into standard (inexpensive) treadmills receiving data from IMU sensors, providing an affordable alternative to instrumented treadmills while allowing broader data collection.

As an application scenario (where such near real-time prediction on embedded devices would be useful), we consider a coach making suggestions to a group of (e.g., a dozen) distance runners training on treadmills. Given \textsc{ser}'s fast inference, the coach can simultaneously monitor GRFs and biomechanical variables (e.g, on a tablet) for the entire group of athletes and adjust their running styles based on recent steps (estimation is available after every step with SER as opposed to after a few minutes with LSTM). For example, the coach can identify fatigue from GRFs \cite{bazuelo2018effect} and adjust the athlete's pace to maintain good running form as a means to prevent injuries.

To quantitatively motivate such applications, we deployed the GRF estimation methods presented in our paper on a Samsung S20 smartphone (Exynos CPU, 2.73 GHz, using a Mongoose M5 core) and measured latency and memory usage during inference for 10 steps (this is the minimum possible number of steps for the \textsc{lstm} models). On average, \textsc{lstm} inference required 14.9 seconds, while \textsc{ser} required only 0.3~ms and \textsc{knn} required 36~ms; memory usage was 180~MB for \textsc{lstm}, 170 MB for \textsc{knn} for \textsc{others} and \textsc{everyone} models and 9 MB for \textsc{personal}, and only 4 MB for \textsc{ser}.

When integration on mobile or embedded devices is not required, limited networking and privacy requirements also pose challenges to the use of cloud resources needed by \textscp{lstm}. Even when using cloud resources is an option, their cost would be orders of magnitude lower for \textsc{knn} and \textsc{ser}.

\subsection{Artifacts in Estimated Waveforms}

We observe that each machine learning method can result in different anomalies in the estimated GRFs, which are significant for its evaluation by domain experts.

\begin{figure*}[ht]
    \centering
     \begin{subfigure}[b]{0.49\linewidth}
         \centering
         \includegraphics[width=\linewidth]{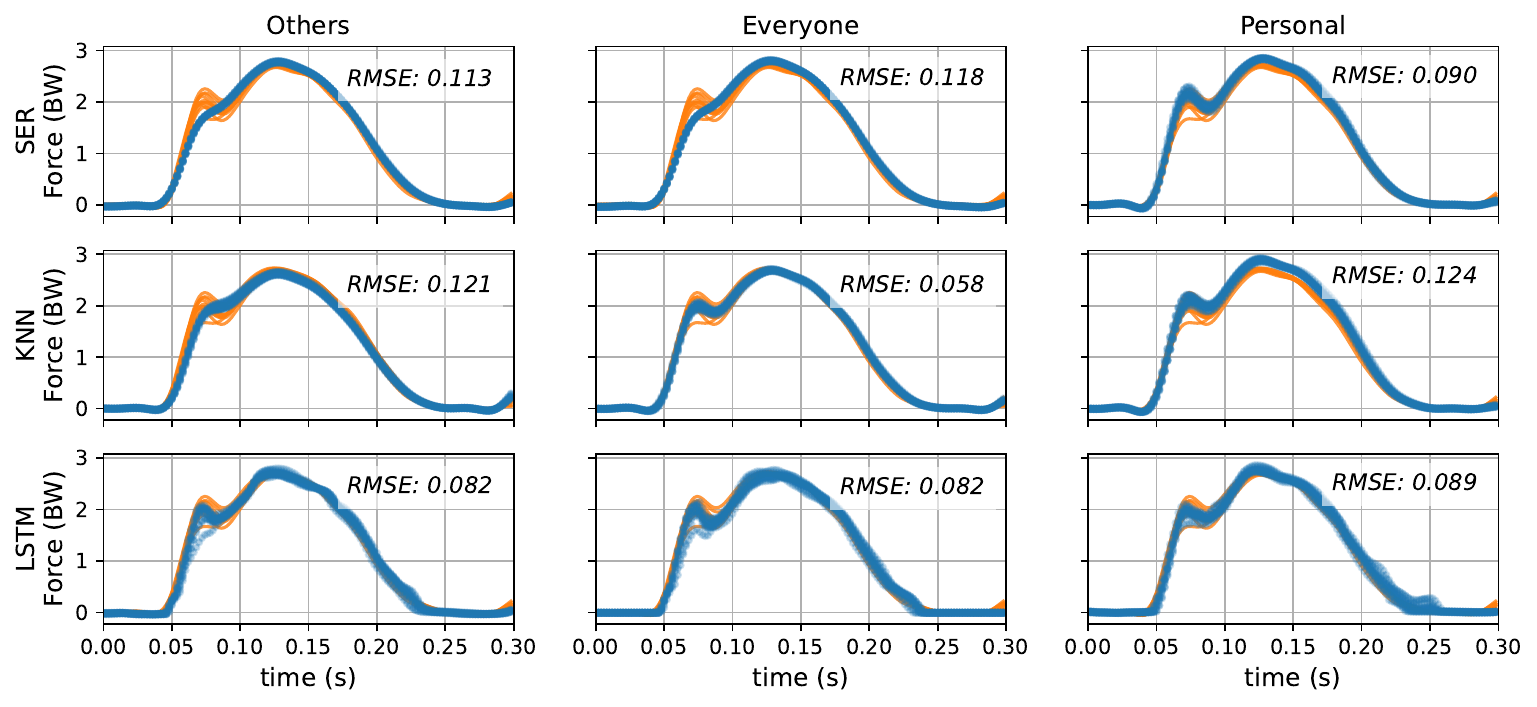}
         \caption{Athlete 1 (Major Impact Peak), Signals \textsc{all}}\label{fig:artifacts1-all}
     \end{subfigure}
     \hfill
     \begin{subfigure}[b]{0.49\linewidth}
         \centering
         \includegraphics[width=\linewidth]{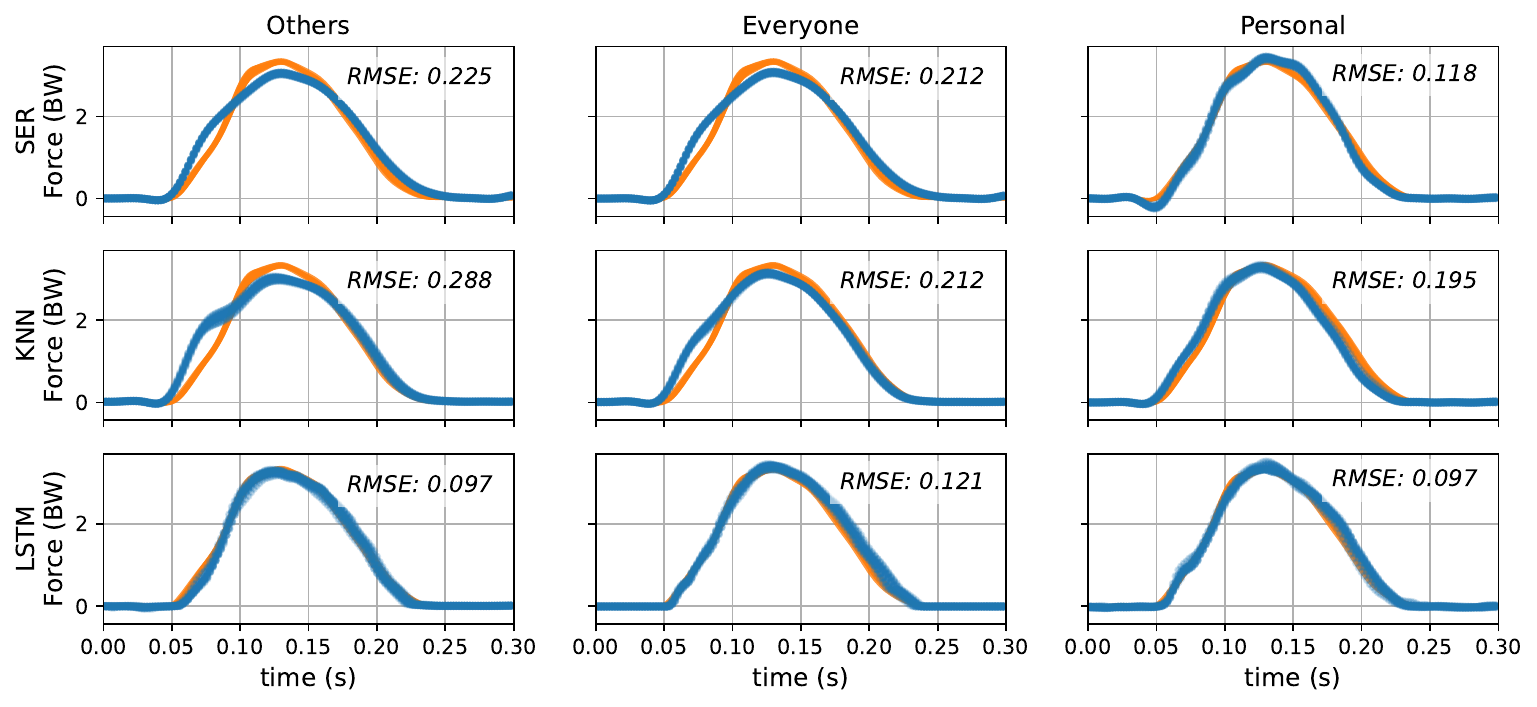}
         \caption{Athlete 2 (No Impact Peak), Signals \textsc{all}}\label{fig:artifacts2-all}
     \end{subfigure}
     \hfill
     \begin{subfigure}[b]{0.49\linewidth}
         \centering
         \includegraphics[width=\linewidth]{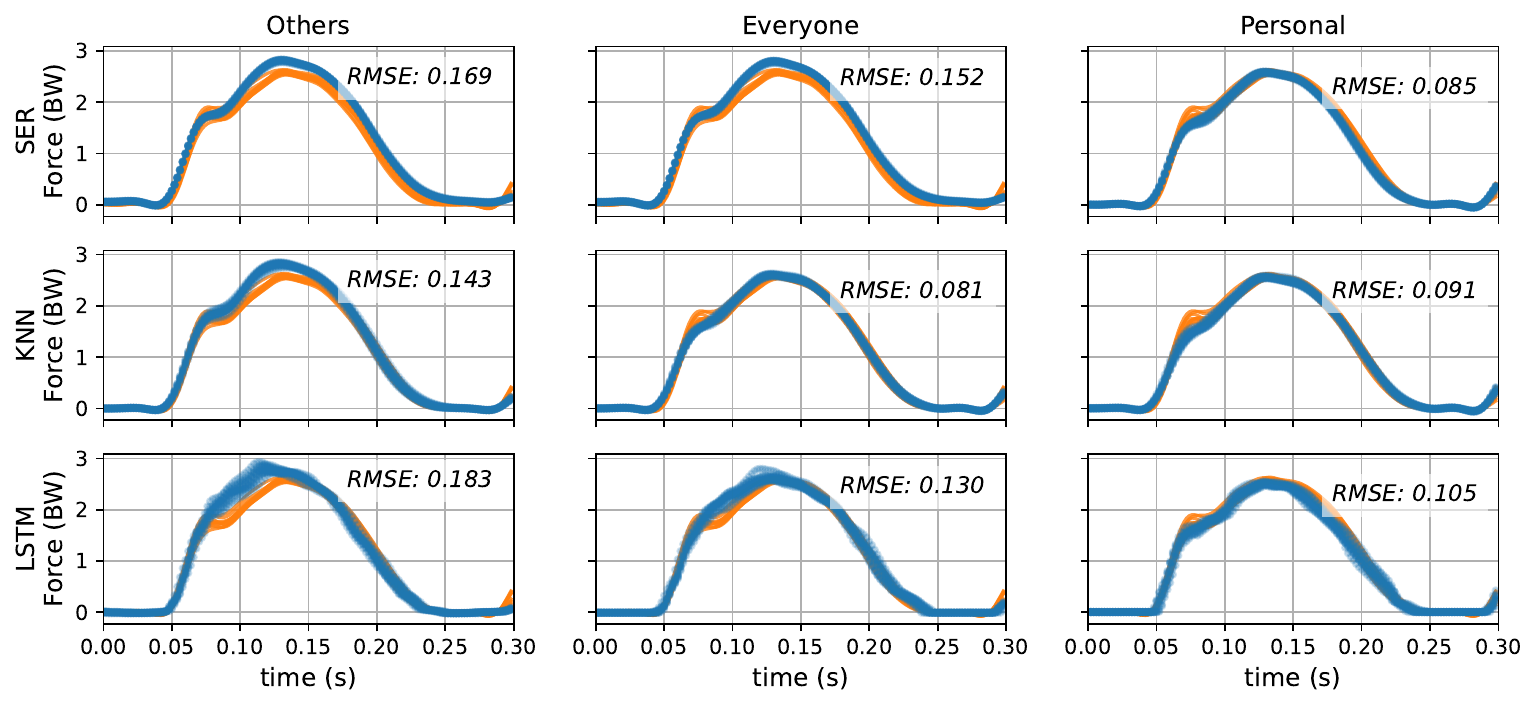}
         \caption{Athlete 3 (Minor Impact Peak), Signals \textsc{all}}\label{fig:artifacts3-all}
     \end{subfigure}
     \hfill
     \begin{subfigure}[b]{0.49\linewidth}
        \centering
        \includegraphics[width=\linewidth]{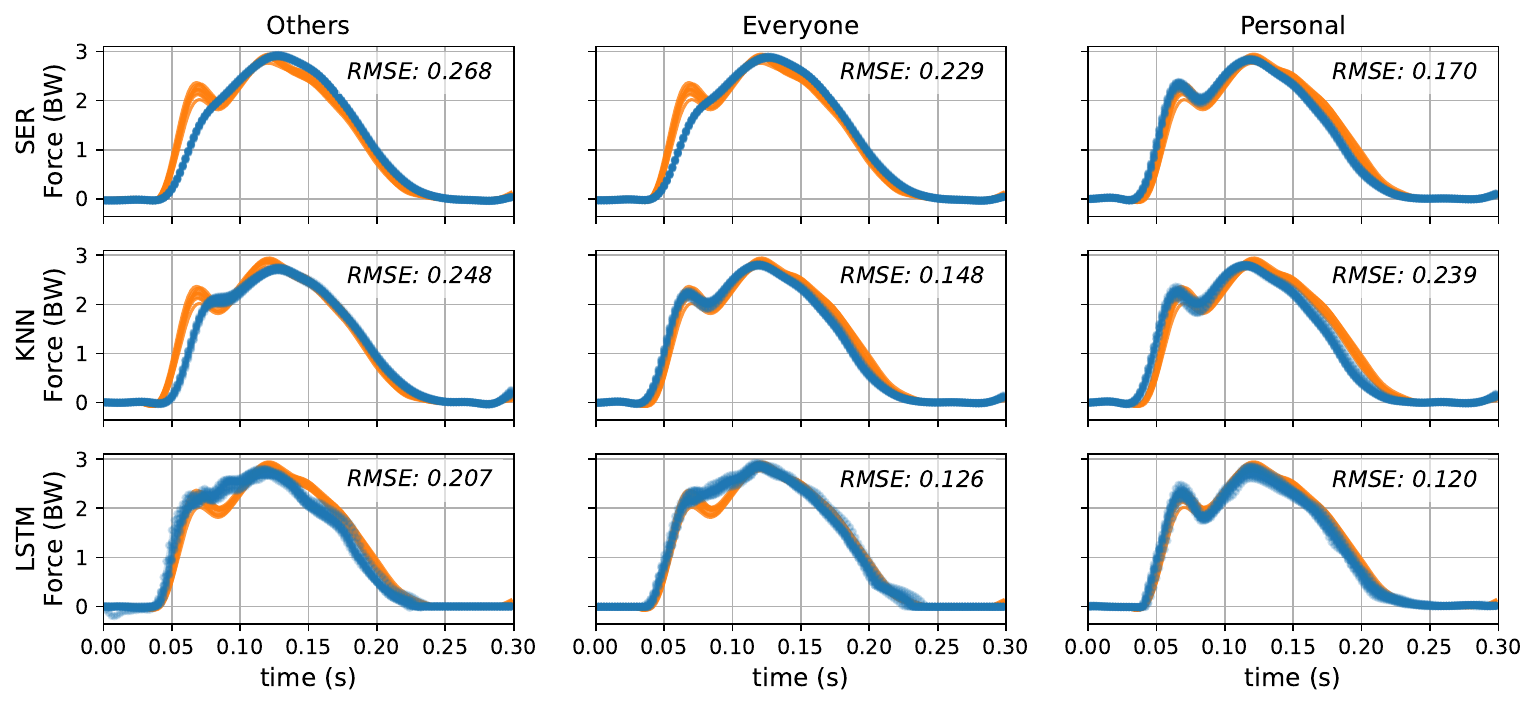}
        \caption{Athlete 4 (Major Impact Peak), Signals \textsc{all}}\label{fig:artifacts4-all}
    \end{subfigure}
    \hfill
     \begin{subfigure}[b]{0.49\linewidth}
         \centering
         \includegraphics[width=\linewidth]{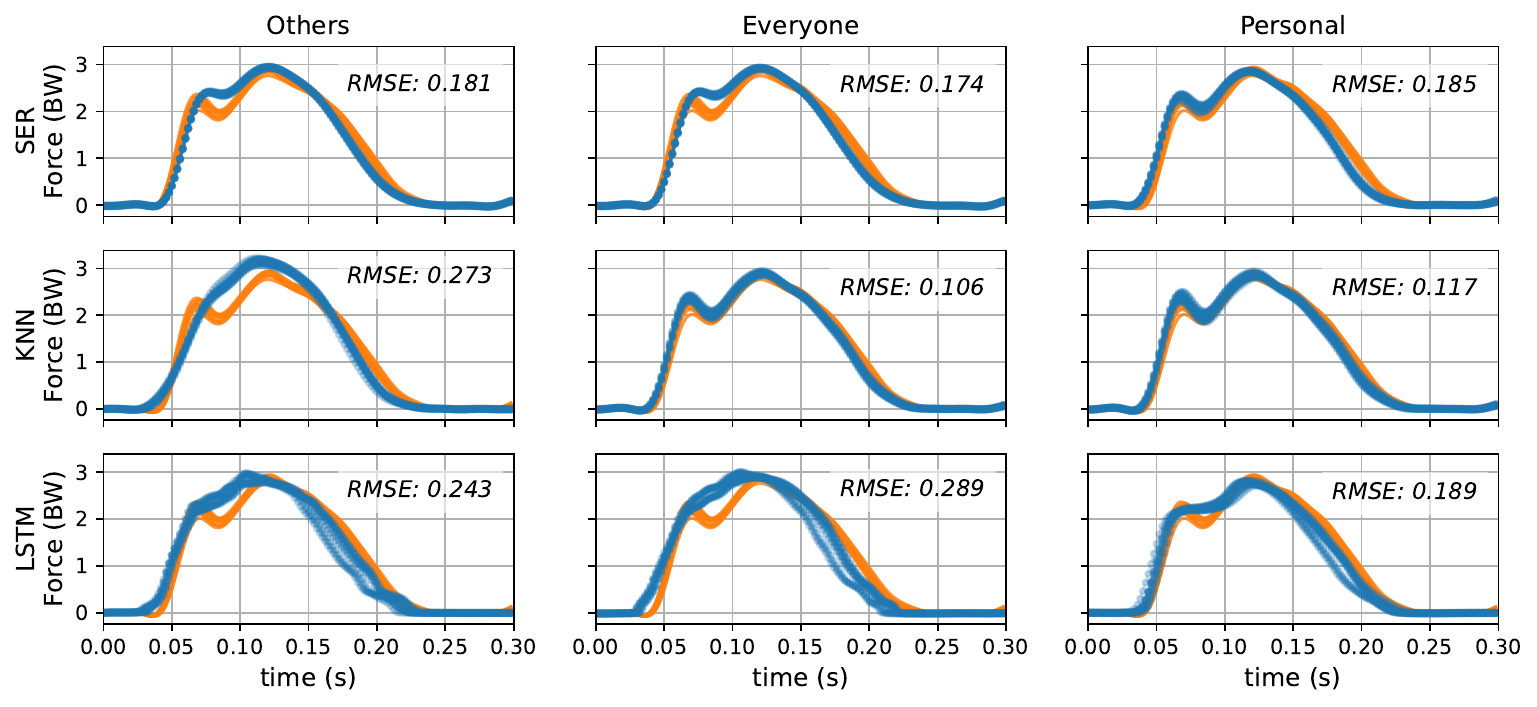}
         \caption{Athlete 4 (Major Impact Peak), Signals \textsc{sac/acc}}\label{fig:artifacts4-sacacc}
     \end{subfigure}
     \hfill
     \begin{subfigure}[b]{0.49\linewidth}
         \centering
         \includegraphics[width=\linewidth]{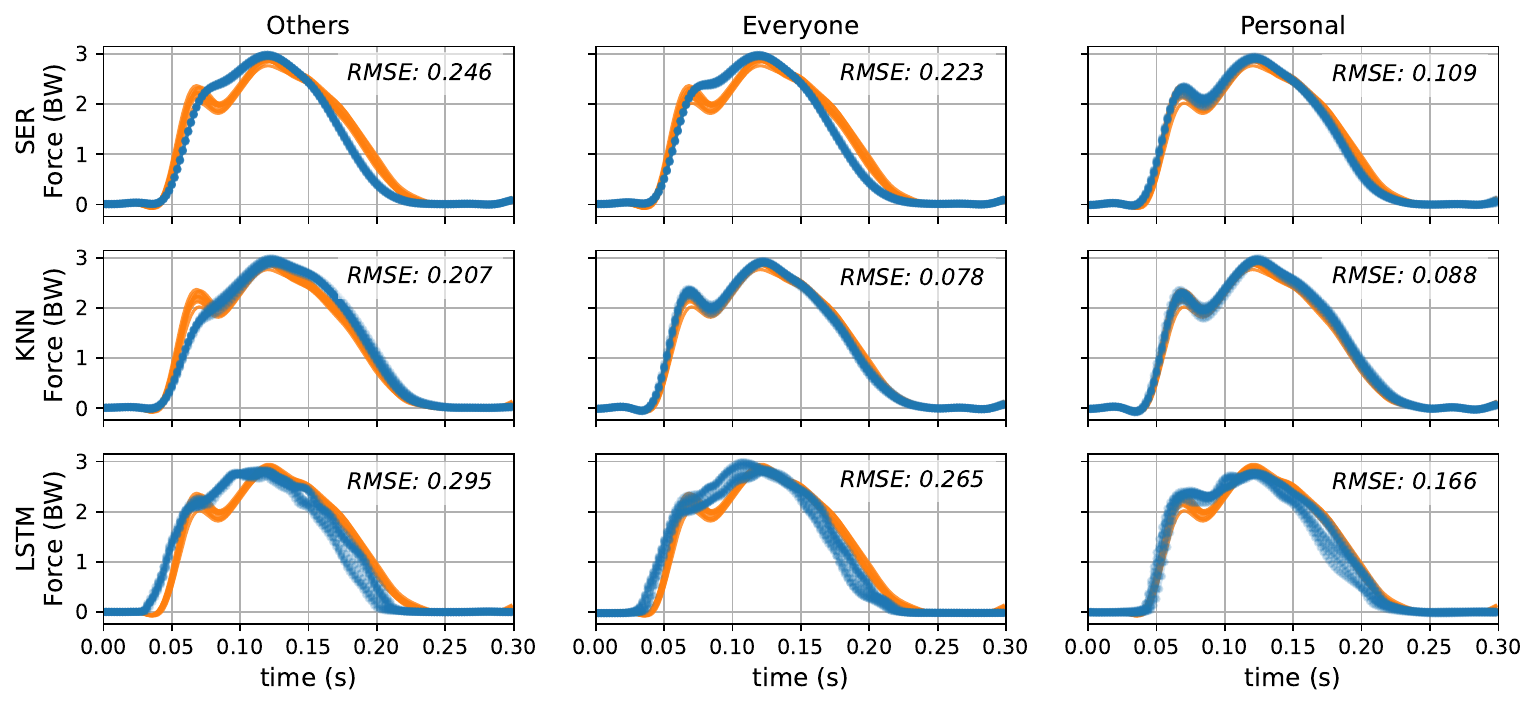}
         \caption{Athlete 4 (Major Impact Peak), Signals \textsc{acc3d}}\label{fig:artifacts4-acc3d}
     \end{subfigure}
     \hfill
    \caption{Comparing estimations (blue) and measurements (orange) of bodyweight normalized vertical GRF $g_z(t)$ for selected athletes and combinations of input signals. In each subgraph, we present the different scenarios (columns \textsc{others}, \textsc{everyone}, \textsc{personal}) and machine learning methods (rows \textsc{ser}, \textsc{knn}, \textsc{lstm}).}\label{fig:prediction_visual}
\end{figure*}

For example, the estimations in \cref{fig:artifacts1-all} are for a runner who initiates ground contact with their heel (rearfoot strike) which leads to a pronounced impact peak in the vertical component of the GRF.
%
In the scenario \textsc{others}, only \textsc{lstm} estimates a pronounced impact peak, while \textsc{knn} and \textsc{ser} estimate smoother waveforms due to the averaging of data from midfoot and forefoot runners (initiating ground contact with their mid or forefoot) with less pronounced impact peaks. Both \textsc{lstm} and \textsc{knn} are able to use data of rearfoot strike runners in the scenarios \textsc{everyone} and \textsc{personal}, while \textsc{ser} accurately estimates this feature only in the scenario \textsc{personal}.

In \cref{fig:artifacts2-all}, we report the GRF of a runner who initiates contact with the front of their foot (forefoot strike) which results in no impact peak in the vertical GRF. In the scenarios \textsc{others} and \textsc{everyone}, \textsc{knn} and \textsc{ser} introduce a change in slope not present in the measured GRF; in contrast, \textsc{lstm} introduces a change in slope in the scenario \textsc{personal}. When athletes have a minor impact peak as in \cref{fig:artifacts3-all}, \textsc{knn} and \textsc{ser} tend to produce more representative waveforms, in all scenarios.

Finally, the athlete of \cref{fig:artifacts4-all} also has a pronounced impact peak. 
While \textsc{knn} and \textsc{ser} estimate a smoother waveform resulting from the average of these patterns, \textsc{lstm} tries to capture both the impact peak and also the active peak, resulting in inaccurate waveforms in the \textsc{others} and \textsc{everyone} scenarios.
The same phenomenon is observed when using only acceleration signals at the sacrum (\cref{fig:artifacts4-sacacc}); when using multidimensional acceleration signals (\textsc{acc3d}), similar anomalies are produced by \textsc{ser} (cases \textsc{others} and \textsc{everyone}).

\subsection{Estimation of Discrete Biomechanical Variables}
\label{sec:gait_metrics_accuracy}

\begin{table*}[tb]\centering
    \ra{1.3}
    \begin{tabular}{@{}r*{13}{S[table-format=2.1]}@{}}\toprule
        & \multicolumn{3}{c}{Scenario \textsc{others}} && \multicolumn{4}{c}{Scenario \textsc{everyone}} && \multicolumn{3}{c}{Scenario \textsc{personal}}&\\\cmidrule{2-4}\cmidrule{6-9}\cmidrule{11-13}
        Biomechanical Variable & \textsc{ser} & \textsc{knn} & \textsc{lstm} && \textsc{ser} & \textsc{knn} & \textsc{lstm} & \shortstack{\textsc{lstm}\\\textsc{fine-tuned}} && \textsc{ser} & \textsc{knn} & \textsc{lstm} &\\\midrule
        Loading Rate         & 19.4 & 19.6 & \good{11.6}        && 19.4 & \good{6.5} & 13.6 & 8.1 && 7.2 & \good{4.9} & 10.0 \\
        Contact Time         & 9.9 & 10.9 & \good{8.8}          && 12.1 & \good{4.7} & 6.4 & 4.8 && \good{4.3} & \good{4.1} & 4.7 \\
        Braking Time         & \good{3.9} & \good{3.5} & 5.9    && 5.9 & \good{2.0} & 4.3 & 3.4 && 3.1 & \good{1.7} & 4.5 \\
        Braking Percentage   & 9.5 & \good{7.4} & 8.6           && 9.2 & \good{3.9} & 5.1 & 6.0 && \good{4.6} & \good{4.3} & 5.2 \\
        Active Peak          & 5.6 & 4.7 & \good{1.6}           && 5.3 & 2.8 & \good{1.3} & 2.2 && \good{2.9} & \good{2.6} & \good{2.6} \\
        Average Vert. Force  & 5.9 & 4.9 & \good{3.2}           && 6.2 & \good{2.5} & 3.1 & 3.2 && \good{2.4} & \good{2.2} & 2.8 \\
        Net Vertical Impulse & 10.5 & 8.3 & \good{5.6}          && 10.6 & \good{4.3} & 5.6 & 6.1 && \good{4.1} & \good{4.3} & 8.4 \\
        A/P Velocity Change  & 12.8 & \good{11.1} & \good{11.4} && 10.0 & \good{5.4} & 9.3 & 6.1 && 7.3 & \good{6.4} & 8.7 \\
    \bottomrule
    \end{tabular}
    \caption{MAPE (\%) of discrete biomechanical variables estimated from \textsc{all} input signals for different scenarios and machine learning methods (results highlighted in blue are optimal or less than 0.5\% from optimal for a scenario and biomechanical variable)}
    \label{tab:gait_metrics_results}
\end{table*}

Finally, we use the estimated GRF waveform to evaluate discrete biomechanical variables (defined precisely in Appendix~\ref{app:gait_metrics}), which are of interest to domain experts for the detection of running anomalies that may lead to stress-responses in bone and soft tissue. MAPE values with respect to the measured GRFs are reported in \cref{tab:gait_metrics_results} for GRF estimations using \textsc{all} input signals, for different scenarios and machine learning methods.

In the scenario \textsc{others}, \textsc{lstm} achieves substantially lower MAPE for most biomechanical variables (except braking time and braking percentage), in line with the lower RMSE and rRMSE values of the estimated GRF waveform (\cref{tab:rmse_z,tab:rrmse_z}, row \textsc{all}).
In contrast, in the scenario \textsc{everyone}, \textsc{lstm} achieves substantially higher MAPE than \textsc{knn}, despite their similar RMSE and rRMSE values in \cref{tab:rmse_z,tab:rrmse_z}. Even with fine-tuning, where \textsc{lstm} has a marginally lower RMSE, \textsc{lstm} still achieves higher MAPE than \textsc{knn} except for one variable (i.e.: Active Peak).

In the scenario \textsc{personal}, \textsc{knn} and \textsc{ser} achieve substantially lower MAPE than \textsc{lstm} for several biomechanical variables, e.g., loading rate and net vertical impulse, as expected.
Notably, while the best RMSE and rRMSE achieved by \textsc{knn} in the scenario \textsc{personal} (\cref{tab:rmse_z,tab:rrmse_z}, row \textsc{all}) are worse than those of the scenario \textsc{everyone}, its MAPE values are substantially lower for most biomechanical variables; unexpected MAPE reductions are observed in the scenario \textsc{personal} also for \textsc{lstm} (e.g., for loading rate and contact time). In general, similarly to related work \cite{Alcantara22}, MAPE values are significantly higher for biomechanical variables that depend on sensitive features of the GRF waveform, e.g., the loading rate, which depends on rate of change of the vertical GRF.

This analysis shows that RMSE and rRMSE of the estimated GRFs are not always useful estimates of the error of the derived biomechanical variables (which depend on specific features of the GRF waveform), and that \emph{personal data is especially useful when biomechanical variables are of interest}. The presence of various types of artifacts in the estimations from \textsc{others}' data, despite achieving a low RMSE, highlights the potential for further improvements. 

\subsection{Provenance}
\label{sec:provenance}
Given the linear nature of \textsc{ser}, it provides greater interpretability of GRF estimation than a highly non-linear model such as \textsc{lstm}.
Moreover, \textsc{ser} also allows tracing each part of the GRF estimation back to the training IMU/GRF data through the model parameters (\textsc{knn} does this for the entire GRF estimation).
This is often referred to as provenance, e.g., as in \cite{yan2016fine}.
Ability to determine provenance is useful for both biomechanics research (to study the relationship between IMU data and GRFs) and practical applications (e.g., to detect which training data is causing model deterioration once it has been deployed in the field for a while). We are not aware of any provenance results in the literature for \textsc{lstm} models, possibly because of the complex relationship between inputs and outputs due to long-term and short-term memory as well as the stochastic nature of the training process.

We also note that provenance is different from sensitivity analysis of the model outputs around specific inputs during inference, e.g., as in \cite{zhou2016learning}. Although such sensitivity analysis could potentially be adapted to \textscp{lstm} (no such work exists to our knowledge), it does not provide a relationship between training data and inference outputs.

\section{Related Work\label{sec:related}}

\begin{table*}[ht]
\renewcommand{\arraystretch}{1.8}
\begin{tabular}{@{}p{0.8cm}p{3.3cm}p{2cm}p{2cm}p{2.3cm}ccccccc@{}}
\toprule
\multirow{2}{*}{Paper}          & \multirow{2}{*}{GRF Measurement}                                  & \multirow{2}{*}{Sensors Type}                                 & \multirow{2}{*}{Sensor Locations}                  & \multirow{2}{*}{Estimation Model} & \multicolumn{3}{c}{RMSE (BW)} && \multicolumn{3}{c}{rRMSE (\%)} \\
\cmidrule{6-8}\cmidrule{10-12}
                                &                                                                   &                                                               &                                                    &                                   & $x$      & $y$      & $z$     && $x$      & $y$      & $z$      \\
\midrule
\cite{Alcantara22}              & Instrumented treadmill, \newline multiple speeds and slopes       & IMU (acceleration)                                            & Sacrum                                             & LSTM                              & ---       & ---       & 0.16    && ---       & ---       & 6.4      \\
\cite{TedescoPKJBHO20}          & In-sole sensors, \newline multiple speeds                         & IMU (acceleration)                                            & Left/right shanks                                  & MLP                               & ---       & ---       & 0.15    && ---       & ---       & ---       \\
\cite{Dorschky20}               & Force plate, multiple speeds                                      & IMU (acceleration, angular velocity)                          & Sacrum, right\newline thigh, shank, foot                   & Simulation + \newline CNN                  & ---       & ---       & ---      && ---       & 2        & 6        \\
\cite{Johnson21}                & Instrumented force plates \newline and treadmills                 & Video, IMU\newline (acceleration)                             & Sacrum, left/right shanks, thighs                  & CNN                               & ---       & ---       & ---      && 21.6     & 17.1     & 13.9     \\
\cite{Wouda18}                  & Instrumented treadmill,\newline multiple speeds                   & Video, IMU\newline (acceleration)                             & Sacrum, left/right shanks                          & MLP                               & ---       & ---       & 0.27    && ---       & ---       & ---       \\
\cite{scheltinga2023estimating} & Instrumented treadmill, \newline heel strike runners                                            & IMU\newline (acceleration)                                                           & Feet, (proximal) tibias, thighs,\newline pelvis, and trunk & Physical model               & 0.05     & 0.07     & 0.18    && 10.8     & 7.8      & 6.6      \\
\midrule
\multirow{3}{*}{Ours}           & \multirow{3}{*}{\shortstack[l]{Instrumented treadmill, \\ multiple speeds}} & \multirow{3}{*}{\shortstack[l]{IMU\\(acceleration, \\angular velocity)}} & \multirow{3}{*}{\shortstack[l]{Sacrum, left/right\\shanks}}         & SER (\textsc{personal})                    & ---       & 0.05    & 0.13    && ---       & 5.0      & 4.2      \\
                                &                                                                   &                                                               &                                                    & KNN (\textsc{everyone})                    & ---       & 0.05    & 0.12    && ---       & 4.6      & 3.9      \\
                                &                                                                   &                                                               &                                                    & LSTM (\textsc{others})                     & ---       & 0.06    & 0.13    && ---       & 7.1      & 4.2     
\\\bottomrule
\end{tabular}
\caption{Comparison with state-of-the-art studies on GRF estimation. Results are reported for each of our methods (\textsc{ser}, \textsc{knn}, \textsc{lstm}) when using \textsc{all} input signals in their best scenarios of application (\textsc{personal}, \textsc{everyone}, \textsc{others}, respectively); other studies consider the \textsc{others} scenario.}
\label{tab:related}
\end{table*}

Similarly to the \textsc{ser} method proposed in this paper, the approach of \cite{pogson2020neural} uses the general idea of transduction \cite{Cortes05} between the embeddings of input IMU data and output GRF data. We observe the following critical differences between \textsc{ser} and \cite{pogson2020neural}:
\begin{itemize}
\item \textsc{ser} uses a different organization of the training data, where the running steps and their IMU and GRF signals are split into \emph{batches}, as shown in \cref{fig:accuracy_vs_steps}. Batch size is a critical hyperparameter to consider intra-step interactions (optimized with a validation set), while using a single step without differentiating left and right foot (as in \cite{pogson2020neural})  results in higher estimation error. This  suggests that the biomechanics of one leg may be different from that of the other.
\item \textsc{ser} uses SVD instead of PCA (used in~\cite{pogson2020neural}), i.e., it does not normalize each input variable of IMU or GRF time series across the entire dataset. This difference allows us to preserve the patterns of the step signals over time.
\item \textsc{ser} uses least squares regression to predict the output embedding instead of neural networks (used in~\cite{pogson2020neural}); in our experiments, neural networks (with up to 5 layers and 100 neurons per layer) resulted in higher estimation error and slower training, due to the limited available data and additional hyperparameter optimization.
For instance, using signal input \textsc{all}, neural networks predicting output embedding achieve RMSE of $0.220$~BW for \textsc{others} ($0.023$~BW higher than least squares regression), $0.189$~BW for \textsc{everyone} ($0.002$~BW higher), and $0.190$~BW for \textsc{personal} scenario ($0.060$~BW higher).
\end{itemize}
Notably, \textsc{ser} performs better than \cite{pogson2020neural} when applied in the same scenario (\textsc{others}), with much lower training times.

Other related works consider different types of model inputs or outputs.
In \cite{Alcantara22,donahue2023estimation}, IMU signals are used to estimate only the vertical GRF, while our study estimates also the anterior-posterior GRF. This enables the derivation of biomechanical variables such as anterior-posterior loading rate (as emphasized in~\cite{johnson2021relationships}), in addition to anterior-posterior braking time and A/P velocity change (evaluated in \cref{tab:gait_metrics_results}).
A higher number of IMU sensors is used in \cite{scheltinga2023estimating} and \cite{wouda2018estimation} (8 and 17 IMUs, respectively). In \cite{scheltinga2023estimating}, a physical model shows performance superior to machine learning approaches in estimating vertical GRF of rearfoot strike runners.
The use of motion capture cameras is explored in\cite{komaris2019predicting}, while \cite{honert2022estimating} uses insole plantar pressure sensors.
Notably, our study offers  insights into the error reductions obtained from different combinations of input IMU signals.

In \cref{tab:related}, we report a summary of state-of-the-art studies on GRF estimation, their settings, and  estimation errors.

\section{Conclusions\label{sec:conclusions}}

GRF waveforms measured during foot contact and their derived biomechanical variables can be accurately estimated from acceleration and angular velocity signals collected using wearable IMU sensors.
To this end, depending on the training data and input signals, simple machine learning methods such as SER and KNN are similarly accurate or more accurate than LSTM neural networks, using fewer computation resources or energy and with much faster inference time on edge devices, training times and hyperparameter optimization, as illustrated by our evaluation.

Notably, SER and KNN produce more accurate estimations of the GRF waveform when personal training data (i.e., GRF and IMU measurements for an athlete) are available; in this case, the error of the estimated biomechanical variables is greatly improved with respect to LSTM neural networks.
We also observed that all machine learning methods benefit from the use of both acceleration and angular velocity, and from the use of all components of the sacrum acceleration (instead of its L2 norm).

In future work, we plan to evaluate the use of estimated GRF waveforms and biomechanical variables for the detection of running anomalies leading to injuries.
We are also interested in a deeper exploration of provenance to understand which training data determined different events and characteristics of the estimated GRF, which we hope will also aid in anomaly detection.

\bibliographystyle{elsarticle-num}
\bibliography{main}

\begin{thebibliography}{10}
\expandafter\ifx\csname url\endcsname\relax
  \def\url#1{\texttt{#1}}\fi
\expandafter\ifx\csname urlprefix\endcsname\relax\def\urlprefix{URL }\fi
\expandafter\ifx\csname href\endcsname\relax
  \def\href#1#2{#2} \def\path#1{#1}\fi

\bibitem{Munro87}
C.~F. Munro, D.~I. Miller, A.~J. Fuglevand, Ground reaction forces in running: a reexamination, Journal of Biomechanics 20~(2) (1987) 147--155.

\bibitem{Cavanagh80}
P.~R. Cavanagh, M.~A. Lafortune, Ground reaction forces in distance running, Journal of Biomechanics 13~(5) (1980) 397--406.

\bibitem{James78}
S.~L. James, B.~T. Bates, L.~R. Osternig, Injuries to runners, The American Journal of Sports Medicine 6~(2) (1978) 40--50.

\bibitem{Hreljac04}
A.~Hreljac, Impact and overuse injuries in runners, Medicine and Science in Sports and Exercise 36~(5) (2004) 845--849.

\bibitem{Napier18}
C.~Napier, C.~MacLean, J.~Maurer, J.~Taunton, M.~Hunt, Kinetic risk factors of running-related injuries in female recreational runners, Scandinavian Journal of Medicine \& Science in Sports 28~(10) (2018) 2164--2172.

\bibitem{johnson2020impact}
C.~D. Johnson, A.~S. Tenforde, J.~Outerleys, J.~Reilly, I.~S. Davis, Impact-related ground reaction forces are more strongly associated with some running injuries than others, The American Journal of Sports Medicine 48~(12) (2020) 3072--3080.

\bibitem{vannatta2020biomechanical}
C.~N. Vannatta, B.~L. Heinert, T.~W. Kernozek, Biomechanical risk factors for running-related injury differ by sample population: A systematic review and meta-analysis, Clinical Biomechanics 75 (2020) 104991.

\bibitem{matijevich2019ground}
E.~S. Matijevich, L.~M. Branscombe, L.~R. Scott, K.~E. Zelik, Ground reaction force metrics are not strongly correlated with tibial bone load when running across speeds and slopes: Implications for science, sport and wearable tech, PloS one 14~(1) (2019) e0210000.

\bibitem{rice2024speed}
H.~Rice, M.~Kurz, P.~Mai, L.~Robertz, K.~Bill, T.~R. Derrick, S.~Willwacher, Speed and surface steepness affect internal tibial loading during running, Journal of sport and health science 13~(1) (2024) 118--124.

\bibitem{Bigouette16}
J.~Bigouette, J.~Simon, K.~Liu, C.~L. Docherty, Altered vertical ground reaction forces in participants with chronic ankle instability while running, Journal of Athletic Training 51~(9) (2016) 682--687.

\bibitem{Kiernan18}
D.~Kiernan, D.~A. Hawkins, M.~A. Manoukian, M.~McKallip, L.~Oelsner, C.~F. Caskey, C.~L. Coolbaugh, Accelerometer-based prediction of running injury in national collegiate athletic association track athletes, Journal of Biomechanics 73 (2018) 201--209.

\bibitem{Messier18}
S.~P. Messier, D.~F. Martin, S.~L. Mihalko, E.~Ip, P.~DeVita, D.~W. Cannon, M.~Love, D.~Beringer, S.~Saldana, R.~E. Fellin, et~al., A 2-year prospective cohort study of overuse running injuries: the runners and injury longitudinal study (trails), The American Journal of Sports Medicine 46~(9) (2018) 2211--2221.

\bibitem{Riley08}
P.~O. Riley, J.~Dicharry, J.~Franz, U.~Della~Croce, R.~P. Wilder, D.~C. Kerrigan, A kinematics and kinetic comparison of overground and treadmill running, Medicine \& Science in Sports \& Exercise 40~(6) (2008) 1093--1100.

\bibitem{Kluitenberg12}
B.~Kluitenberg, S.~W. Bredeweg, S.~Zijlstra, W.~Zijlstra, I.~Buist, Comparison of vertical ground reaction forces during overground and treadmill running. a validation study, BMC Musculoskeletal Disorders 13~(1) (2012) 1--8.

\bibitem{Asmussen19}
M.~J. Asmussen, C.~Kaltenbach, K.~Hashlamoun, H.~Shen, S.~Federico, B.~M. Nigg, Force measurements during running on different instrumented treadmills, Journal of Biomechanics 84 (2019) 263--268.

\bibitem{Jacobs15}
D.~A. Jacobs, D.~P. Ferris, Estimation of ground reaction forces and ankle moment with multiple, low-cost sensors, Journal of Neuroengineering and Rehabilitation 12~(1) (2015) 1--12.

\bibitem{mason2023wearables}
R.~Mason, L.~T. Pearson, G.~Barry, F.~Young, O.~Lennon, A.~Godfrey, S.~Stuart, Wearables for running gait analysis: A systematic review, Sports Medicine 53~(1) (2023) 241--268.

\bibitem{leporace2018prediction}
G.~Leporace, L.~A. Batista, J.~Nadal, Prediction of 3d ground reaction forces during gait based on accelerometer data, Research on Biomedical Engineering 34 (2018) 211--216.

\bibitem{Dorschky20}
E.~Dorschky, M.~Nitschke, C.~F. Martindale, A.~J. van~den Bogert, A.~D. Koelewijn, B.~M. Eskofier, {CNN-Based Estimation of Sagittal Plane Walking and Running Biomechanics From Measured and Simulated Inertial Sensor Data}, Frontiers in Bioengineering and Biotechnology 8 (2020).
\newblock \href {https://doi.org/10.3389/fbioe.2020.00604} {\path{doi:10.3389/fbioe.2020.00604}}.

\bibitem{Johnson21}
W.~R. Johnson, A.~Mian, M.~A. Robinson, J.~Verheul, D.~G. Lloyd, J.~A. Alderson, \href{https://doi.org/10.1109/TBME.2020.3006158}{Multidimensional ground reaction forces and moments from wearable sensor accelerations via deep learning}, {IEEE} Trans. Biomed. Eng. 68~(1) (2021) 289--297.
\newblock \href {https://doi.org/10.1109/TBME.2020.3006158} {\path{doi:10.1109/TBME.2020.3006158}}.
\newline\urlprefix\url{https://doi.org/10.1109/TBME.2020.3006158}

\bibitem{Alcantara21}
R.~S. Alcantara, E.~M. Day, M.~E. Hahn, A.~M. Grabowski, Sacral acceleration can predict whole-body kinetics and stride kinematics across running speeds, PeerJ 9 (2021) e11199.
\newblock \href {https://doi.org/10.7717/peerj.11199} {\path{doi:10.7717/peerj.11199}}.

\bibitem{Alcantara22}
R.~S. Alcantara, W.~B. Edwards, G.~Y. Millet, A.~M. Grabowski, Predicting continuous ground reaction forces from accelerometers during uphill and downhill running: a recurrent neural network solution, PeerJ 10 (2022) e12752.
\newblock \href {https://doi.org/10.7717/peerj.12752} {\path{doi:10.7717/peerj.12752}}.

\bibitem{ren2008whole}
L.~Ren, R.~K. Jones, D.~Howard, Whole body inverse dynamics over a complete gait cycle based only on measured kinematics, Journal of Biomechanics 41~(12) (2008) 2750--2759.

\bibitem{harper2023}
H.~E. Stewart, R.~S. Alcantara, K.~A. Farina, Can ground reaction force variables pre-identify the probability of a musculoskeletal injury in collegiate distance runners, under review (2023).

\bibitem{bair2006prediction}
E.~Bair, T.~Hastie, D.~Paul, R.~Tibshirani, Prediction by supervised principal components, Journal of the American Statistical Association 101~(473) (2006) 119--137.

\bibitem{Cortes05}
C.~Cortes, M.~Mohri, J.~Weston, A general regression technique for learning transductions, in: {ICML} 2005, Vol. 119 of {ACM} International Conference Proceeding Series, {ACM}, 2005, pp. 153--160.
\newblock \href {https://doi.org/10.1145/1102351.1102371} {\path{doi:10.1145/1102351.1102371}}.

\bibitem{venturi2023svd}
S.~Venturi, T.~Casey, Svd perspectives for augmenting deeponet flexibility and interpretability, Computer Methods in Applied Mechanics and Engineering 403 (2023) 115718.

\bibitem{liao2013efficient}
Q.~Liao, Q.~Zhang, Efficient rank-one residue approximation method for graph regularized non-negative matrix factorization, in: Machine Learning and Knowledge Discovery in Databases: European Conference, ECML PKDD 2013, Prague, Czech Republic, September 23-27, 2013, Proceedings, Part II 13, Springer, 2013, pp. 242--255.

\bibitem{nvidia_titan_gpu}
NVIDIA, \href{https://www.nvidia.com/en-us/geforce/products/10series/titan-x-pascal/}{{NVIDIA TITAN X Graphics Card}} (2023).
\newline\urlprefix\url{https://www.nvidia.com/en-us/geforce/products/10series/titan-x-pascal/}

\bibitem{nvidia_titan_energy}
NVIDIA, \href{https://www.nvidia.com/en-us/geforce/graphics-cards/geforce-gtx-titan-x/specifications/}{{GeForce GTX TITAN X | Specifications}} (2023).
\newline\urlprefix\url{https://www.nvidia.com/en-us/geforce/graphics-cards/geforce-gtx-titan-x/specifications/}

\bibitem{intel_cpu_energy}
I.~Corporation, \href{https://www.intel.com/content/www/us/en/products/sku/94189/intel-core-i76800k-processor-15m-cache-up-to-3-60-ghz/specifications.html}{{Intel Core i7-6800k Processor}} (2023).
\newline\urlprefix\url{https://www.intel.com/content/www/us/en/products/sku/94189/intel-core-i76800k-processor-15m-cache-up-to-3-60-ghz/specifications.html}

\bibitem{amz_gpu_renting}
A.~EC2, \href{https://aws.amazon.com/ec2/instance-types/g4/}{Amazon ec2 g4 instances} (2023).
\newline\urlprefix\url{https://aws.amazon.com/ec2/instance-types/g4/}

\bibitem{amz_cpu_renting}
A.~EC2, \href{https://aws.amazon.com/ec2/pricing/on-demand/}{Amazon ec2 on-demand pricing} (2023).
\newline\urlprefix\url{https://aws.amazon.com/ec2/pricing/on-demand/}

\bibitem{bazuelo2018effect}
B.~Bazuelo-Ruiz, J.~V. Dur{\'a}-Gil, N.~Palomares, E.~Medina, S.~Llana-Belloch, Effect of fatigue and gender on kinematics and ground reaction forces variables in recreational runners, PeerJ 6 (2018) e4489.

\bibitem{yan2016fine}
Z.~Yan, V.~Tannen, Z.~G. Ives, Fine-grained provenance for linear algebra operators, in: Proceedings of the 8th USENIX Conference on Theory and Practice of Provenance, 2016, pp. 1--6.

\bibitem{zhou2016learning}
B.~Zhou, A.~Khosla, A.~Lapedriza, A.~Oliva, A.~Torralba, Learning deep features for discriminative localization, in: Proceedings of the IEEE conference on computer vision and pattern recognition, 2016, pp. 2921--2929.

\bibitem{pogson2020neural}
M.~Pogson, J.~Verheul, M.~A. Robinson, J.~Vanrenterghem, P.~Lisboa, A neural network method to predict task-and step-specific ground reaction force magnitudes from trunk accelerations during running activities, Medical Engineering \& Physics 78 (2020) 82--89.

\bibitem{donahue2023estimation}
S.~R. Donahue, M.~E. Hahn, Estimation of gait events and kinetic waveforms with wearable sensors and machine learning when running in an unconstrained environment, Scientific Reports 13~(1) (2023) 2339.

\bibitem{johnson2021relationships}
C.~D. Johnson, J.~Outerleys, I.~S. Davis, Relationships between tibial acceleration and ground reaction force measures in the medial-lateral and anterior-posterior planes, Journal of biomechanics 117 (2021) 110250.

\bibitem{scheltinga2023estimating}
B.~L. Scheltinga, J.~N. Kok, J.~H. Buurke, J.~Reenalda, Estimating 3d ground reaction forces in running using three inertial measurement units, Frontiers in Sports and Active Living 5 (2023) 1176466.

\bibitem{wouda2018estimation}
F.~J. Wouda, M.~Giuberti, G.~Bellusci, E.~Maartens, J.~Reenalda, B.-J.~F. Van~Beijnum, P.~H. Veltink, Estimation of vertical ground reaction forces and sagittal knee kinematics during running using three inertial sensors, Frontiers in Physiology 9 (2018) 218.

\bibitem{komaris2019predicting}
D.-S. Komaris, E.~P{\'e}rez-Valero, L.~Jordan, J.~Barton, L.~Hennessy, B.~O’Flynn, S.~Tedesco, Predicting three-dimensional ground reaction forces in running by using artificial neural networks and lower body kinematics, IEEE Access 7 (2019) 156779--156786.

\bibitem{honert2022estimating}
E.~C. Honert, F.~Hoitz, S.~Blades, S.~R. Nigg, B.~M. Nigg, Estimating running ground reaction forces from plantar pressure during graded running, Sensors 22~(9) (2022) 3338.

\bibitem{yong2018acute}
J.~R. Yong, A.~Silder, K.~L. Montgomery, M.~Fredericson, S.~L. Delp, Acute changes in foot strike pattern and cadence affect running parameters associated with tibial stress fractures, Journal of biomechanics 76 (2018) 1--7.

\end{thebibliography}

\appendices

\section{Aligning Signals from Different Locations by Reference Events}
\label{app:manual_alignment}

To estimate GRFs from IMUs, timestamps and gait events need to match consistently across different signals and locations. For a model that estimates the entire stance in GRFs from signals of an entire stance in IMU signals, training the model would require supplying signals aligned by their corresponding steps. Similarly for models estimating one sample point at a time based on samples of other signals at the same time point, signals at different locations need to synchronize. 
Given acceleration and angular velocity measurements of a IMU sensor are synchronized, we manually aligned data from IMU sensors at different locations and GRF data from the treadmills (which was downsampled to 500 Hz to match the IMU frequency and normalized by the body weight of the athlete).

In our dataset, IMU sensors at different locations (sacrum and shanks) and GRFs measured from force plates are sampled by individual clocks. Although we do not observe any severely non-linear drifts, linear drifts and time delays are significantly affecting the alignment of gait events.

\begin{figure}[ht]
    \centering
    \begin{subfigure}[b]{.45\linewidth}
        \includegraphics[width=\linewidth]{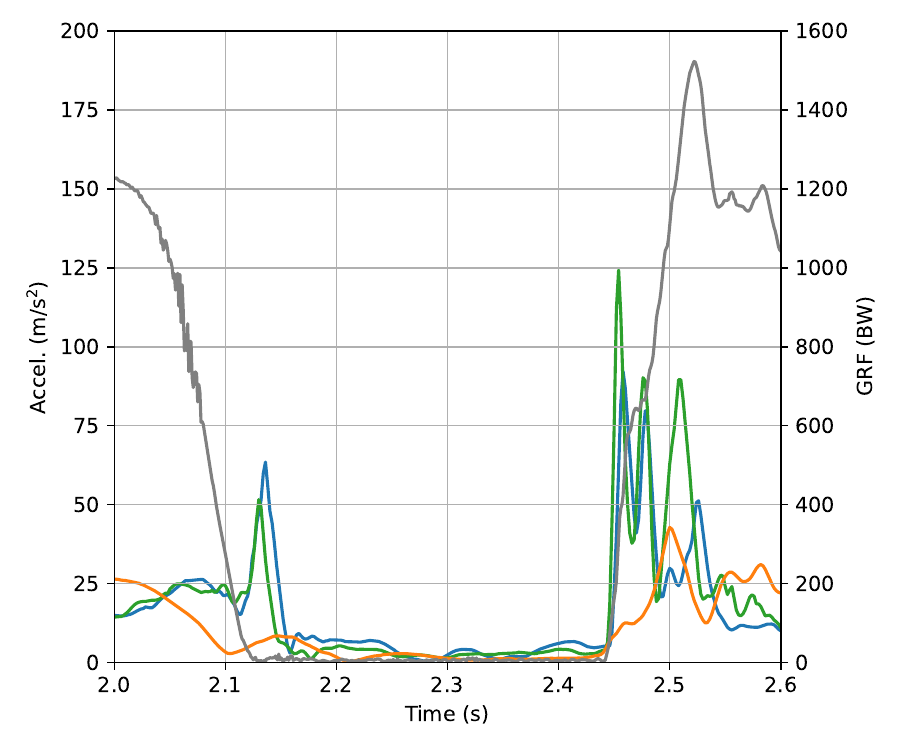}
        \caption{}
        \label{fig:jump_reference}
    \end{subfigure}
    \hfill
    \begin{subfigure}[b]{.45\linewidth}
        \includegraphics[width=\linewidth]{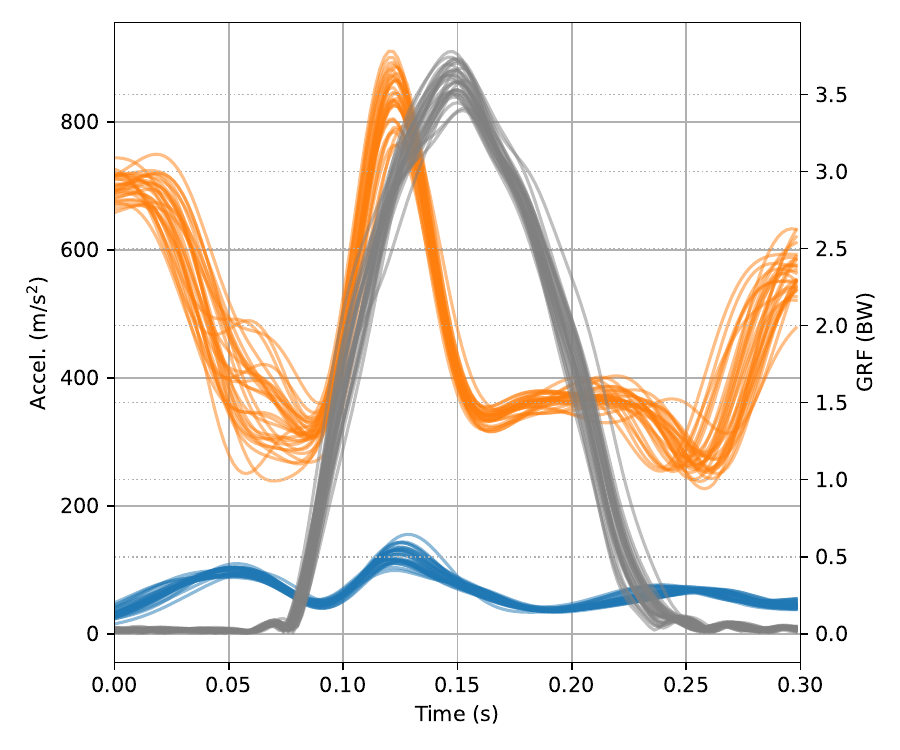}
        \caption{}
        \label{fig:running_steps_overlap_acc}
    \end{subfigure}
    \caption{Examples of overlapping the magnitude of GRFs with the magnitude of acceleration signals from shanks and sacrum. Blue is from left shank, green is from right shank, orange is from sacrum, and grey is from GRFs. \protect\subref{fig:jump_reference}) Aligned jump reference. \protect\subref{fig:running_steps_overlap_acc}) Consecutive left steps under a constant speed.}
\end{figure}

To align all gait events over a series of continuous running steps, the dataset includes reference events where the athlete jumps in-place before and after continuously running on an instrumented treadmill. Wearing all sensors, the athlete jumping in-place creates a unique signal pattern across all sensors with a sharp edge marking the time instance when both feet strike the ground after flight. An example of aligned signals at the jump reference is shown in \cref{fig:jump_reference}. We align both reference events before and after a run by shifting and linearly stretching the signals to correct time delays and linear drifts. We use magnitude of GRFs as the referencing signal and edit IMU signals to align with GRFs.
After aligning both reference events, signals for each running steps between the references are also aligned.

After aligning data from different sensors, each running measurement is automatically split into steps by identifying the maxima of the correlation between the L2~norm of shank acceleration and a reference signal (a triangular signal with 100~ms duration followed by a zero signal of 100~ms, mimicking the patterns observed in acceleration signals). 
Steps of a running measurement are aligned by maximizing their pairwise correlation; then, a fixed delay is applied to all the steps in a measurement to maximize their mean correlation with the reference signal, in order to align them with steps of other measurements (i.e., at different running speeds or for different athletes).
We manually check the alignment of signals for each foot contact by overlapping all steps from each run (an example of such view is shown in \cref{fig:running_steps_overlap_acc}.).
Aligning these signals by the start and end points shows time drifts between signals are only linear and the overlapped view shows gait events within each steps are aligned similarly.
While there is no guarantee for each sample point across all signal types is aligned perfectly, models looking at an entire stance or a windowed signals greater or equal to a stance should have a consistent amount of information.

\begin{table*}[tb]\centering
    \ra{1.3}
    \begin{tabular}{@{}rcccccccccccc@{}}\toprule
        & \multicolumn{3}{c}{Scenario \textsc{others}} && \multicolumn{3}{c}{Scenario \textsc{everyone}} && \multicolumn{3}{c}{Scenario \textsc{personal}}\\\cmidrule{2-4}\cmidrule{6-8}\cmidrule{10-12}
        Input Signals & \textsc{ser} & \textsc{knn} & \textsc{lstm} && \textsc{ser} & \textsc{knn} & \textsc{lstm}  && \textsc{ser} & \textsc{knn} & \textsc{lstm} \\\midrule
        \textsc{all}       & 0.07 &	0.06 &	0.06 &&	0.06 &	0.05 &	0.06 &&	0.05 &	0.05 &	0.06 \\
        \textsc{acc}       & 0.07 &	0.06 &	0.08 &&	0.07 &	0.05 &	0.07 &&	0.05 &	0.05 &	0.07  \\
        \textsc{ang}       & 0.07 &	0.06 &	0.07 &&	0.07 &	0.05 &	0.07 &&	0.05 &	0.05 &	0.07  \\
        \textsc{shank}     & 0.07 &	0.06 &	0.08 &&	0.07 &	0.05 &	0.07 &&	0.05 &	0.05 &	0.07  \\
        \textsc{sacrum}    & 0.08 &	0.07 &	0.12 &&	0.07 &	0.05 &	0.10 &&	0.05 &	0.05 &	0.08  \\
        \textsc{sac/acc3d} & 0.07 &	0.06 &	0.07 &&	0.07 &	0.05 &	0.07 &&	0.05 &	0.05 &	0.08 \\
        \textsc{sac/acc}   &  0.07 &	0.07 &	0.08 &&	0.07 &	0.05 &	0.08 &&	0.05 &	0.05 &	0.11  \\
        \bottomrule
    \end{tabular}
    \caption{RMSE (in body weight units, BW, i.e., N/kg) of anterior/posterior GRF estimations $RMSE(g_y, \hat g_y)$ for different input signals, data scenarios, machine learning methods}\label{tab:rmse_y} 
\end{table*}

\begin{table*}[tb]\centering
    \ra{1.3}
    \begin{tabular}{@{}r*{12}{S[table-format=1.1]}@{}}\toprule
        & \multicolumn{3}{c}{Scenario \textsc{others}} && \multicolumn{3}{c}{Scenario \textsc{everyone}} && \multicolumn{3}{c}{Scenario \textsc{personal}}\\\cmidrule{2-4}\cmidrule{6-8}\cmidrule{10-12}
        Input Signals & \textsc{ser} & \textsc{knn} & \textsc{lstm} && \textsc{ser} & \textsc{knn} & \textsc{lstm} && \textsc{ser} & \textsc{knn} & \textsc{lstm} \\\midrule
        \textsc{all}       & 6.8 &	5.9 &	7.1 && 	6.5 &	4.6 &	7.1 &&	5.0 &	4.9 &	7.2\\
        \textsc{acc}       & 6.9 &	6.2 &	9.2 && 	6.7 &	4.7 &	8.9 &&	4.9 &	4.7 &	8.0  \\
        \textsc{ang}       & 7.0 &	6.2 &	8.9 && 	6.7 &	4.7 &	8.2 &&	5.2 &	5.0 &	8.0 \\
        \textsc{shank}     & 7.1 &	6.4 &	9.2 && 	7.0 &	5.1 &	8.4 &&	5.3 &	4.9 &	8.5 \\
        \textsc{sacrum}    & 7.5 &	7.4 &	16.3 && 	7.3 &	4.9 &	13.1 &&	5.2 &	5.0 &	9.3 \\
        \textsc{sac/acc3d} & 7.1 &	6.4 &	9.0 && 	6.9 &	4.6 &	8.9 &&	4.8 &	4.8 &	9.8 \\
        \textsc{sac/acc}   & 7.2 &	6.8 &	9.6 && 	6.9 &	4.9 &	9.7 &&	5.0 &	5.0 &	15.5  \\
        \bottomrule
    \end{tabular}
    \caption{rRMSE (\%) of anterior/posterior GRF estimations $rRMSE(g_y, \hat g_y)$ for different input signals, data scenarios, machine learning methods}\label{tab:rrmse_y}
\end{table*}

\section{Discrete Biomechanical Variables}
\label{app:gait_metrics}

We consider the following discrete biomechanical variables (or \emph{gait metrics}) from \cite{Munro87,Alcantara22} to evaluate our GRF estimations. Let $g_x, g_y$ and $g_z$ be the components of the GRFs of a step (after a 50 Hz low-pass filter and normalized by body weight, with unit denoted as $BW$); the \emph{start time} $T_s$ (in seconds) is the time when the vertical GRF reaches 50 N, i.e., $T_s = \min \{\, t \mid BW \cdot g_z(t) > 50 \,\}$, while the \emph{end time} $T_e$ is the time when the vertical GRF drops below 50 N, i.e., $T_e = \min \{\,t > T_s \mid BW \cdot g_z(t) < 50 \,\}$.

\begin{itemize}
    \item \emph{Loading Rate} ($BW\cdot s^{-1}$): Average slope of the vertical GRF during the first 25 ms of the stance after reaching the 50 N threshold~\cite{yong2018acute}, i.e.,
    \begin{align*}
        \text{Loading Rate} = \frac{g_z(T_s+0.025)-g_z(T_s)}{0.025}\,.
    \end{align*}
    \item \emph{Contact Time} ($s$): Time during which the vertical GRF is above 50 N, i.e., $T_c = T_e - T_s$.
    \item \emph{Braking Time} ($s$): Time during which the vertical GRF signal is above the threshold and the A/P GRF component is negative, i.e., \[T_b = |\{T_s \leq t \leq T_e \mid g_y(t) < 0\}|\,.\]
    \item \emph{Braking Percentage}: Percentage of contact time spent in braking, i.e., $T_b/T_c$.
    \item \emph{Active Peak} ($BW$): Maximum vertical GRF between 30-100\% of the stance (to exclude the impact peak), i.e., $\max \{\, g_z(t) \,|\, t > T_s + 0.3 T_c \,\}$.
    \item \emph{Average Vertical Force} $(BW)$: Average value of the vertical GRF, i.e., $\frac{1}{T_c}\int_{T_s}^{T_e} g_z(t) \,dt$.
    \item \emph{Net Vertical Impulse} ($BW\cdot s$): Area under the vertical GRF  reduced by the body weight unit, i.e., $\big(\int_{T_s}^{T_e} g_z(t)\,dt \big) - 1$.
    \item \emph{A/P Velocity Change} $(m \cdot s^{-1})$: Change in velocity along the A/P force direction, i.e., $9.81 \cdot(\textit{A/P Impulse})$, where the \emph{A/P Impulse} ($BW\cdot s$) is the area between the A/P GRF component and the zero line, i.e., $\int_{T_s}^{T_e} g_y(t) \,dt$.
\end{itemize}

\section{RMSE and Relative RMSE of\\Anterior/Posterior GRF}
This appendix reports the estimation errors for anterior-posterior GRF ($y$~direction) in \cref{tab:rmse_y,tab:rrmse_y}, while estimation errors for the vertical GRF ($z$~direction) are reported in \cref{tab:rmse_z,tab:rrmse_z} of the main text. The GRFs along both vertical and anterior-posterior  directions are used to compute biomechanical variables in \cref{tab:gait_metrics_results}.

\FloatBarrier
\section*{Acknowledgement}

We would like to acknowledge the contribution of Qinming Zhang in data processing.
This project was supported in part by the Pac-12 Student-Athlete Health \& Well-Being Grant Program (\texttt{\#3-03\_PAC-12-Oregon-Hahn-17-02}). This work was supported in part through the Viterbi School of Engineering. This work was supported in part by a grant from Novartis.

\end{document}